\documentclass[journal]{IEEEtran}

\usepackage[pdftex]{graphicx}
\ifCLASSINFOpdf
\else
\fi
\usepackage{xcolor}

\usepackage{amsmath}
\usepackage{amsfonts}
\usepackage{soul}
\usepackage{array}
\usepackage[caption=false,font=normalsize,labelfont=sf,textfont=sf]{subfig}
\begin{document}
\bibliographystyle{unsrt}
\title{Understanding Neural Networks with Logarithm Determinant Entropy Estimator
}
% \title{LogDet: Making Informative Analysis Reliable in Deep Neural Networks}

\author{\IEEEauthorblockN{Zhanghao Zhouyin\IEEEauthorrefmark{1},
and Ding Liu\IEEEauthorrefmark{1},
}
        
\IEEEauthorblockA{\IEEEauthorrefmark{1}School of Computer Science and Technology, Tiangong University, Tianjin 300387, China}
% <-this % stops an unwanted space
\thanks{Manuscript received ; revised ** **, .
Corresponding author: Ding Liu (email: liuding@tiangong.edu.cn).}}

\maketitle

\begin{abstract}
Understanding the informative behaviour of deep neural networks is challenged by misused estimators and the complexity of network structure, which leads to inconsistent observations and diversified interpretation. Here we propose the LogDet estimator -- a reliable matrix-based entropy estimator that approximates Shannon differential entropy. We construct informative measurements based on LogDet estimator, verify our method with comparable experiments and utilize it to analyse neural network behaviour. Our results demonstrate the LogDet estimator overcomes the drawbacks that emerge from highly diverse and degenerated distribution thus is reliable to estimate entropy in neural networks. The Network analysis results also find a functional distinction between shallow and deeper layers, which can help understand the compression phenomenon in the Information bottleneck theory of neural networks.
\end{abstract}


% \fi
\IEEEpeerreviewmaketitle

\section{Introduction}
\IEEEPARstart{U}{sing} 
Information Bottleneck theory (IB) \cite{tishby2000information}\cite{shwartz2017opening} to analyze neural network's behaviour has been found applicable in various domain. According to IB Theory, the learning process can be characterized as finding an optimal representation that captures most target information, while having the least dependency on the original signal. However, recent analytical research of neural network's informative behaviour achieves highly diversified results, where both works that approve and oppose are reported, hardly reaching a certain consensus. Criticisms state that current information estimators are vulnerable to different model and saturate in high dimension feature space\cite{geiger2020information}\cite{cheng2019utilizing}\cite{saxe2019information}, some are accused fallacious measure the sample geometry, rather than informative behaviour\cite{goldfeld2018estimating}\cite{schiemer2019revisiting}. These debates and accuses discredit analytic results and attach importance to revisit informative estimators. Facing the challenge, this work provides a reliable matrix-based Logarithm Determinant (LogDet) entropy estimator. Rather than counting numbers of resembling samples, it recognizes feature-wise dependency and is derived directly from Shannon differential entropy. Our experiments reveal LogDet estimator is robust to noise and random perturbation in any dimension of feature space.

Information Bottleneck Theory (IB), as an extension of minimal sufficient statistics and rate distortion theory \cite{tishby2000information}, describe the learning process as extracting minimal representation $T$ from input $X$, where it contains the most relevant information of target $Y$. Thus, the objective of IB is described with mutual information as:
\begin{equation*}
    \min I(X;T)-\beta I(T;Y).
\end{equation*}
Further, when applied in neural networks, it propose that the learning process can be characterized into two distinct phases by tracking $I(X;T)$ and $I(T;Y)$: \textbf{(1)} \textbf{Fitting} where Networks extract information from input with rising $I(X;T)$ and $I(T;Y)$; \textbf{(2)} \textbf{Compression} that models reduce redundancies and their dependency on input $X$, as a decrease of $I(X;T)$\cite{shwartz2017opening}. Therefore, by tracking mutual information of NNs, we clearly understand the training behaviour from an informative aspect and therefore “Opening the Black Box of Neural Networks”\cite{geiger2020information}. However, some theoretical problems still exist in entropy estimation. To any neural network, each layer output $T_i$ is determined with the sure input $X$. When considering $T$ and $S$ are both discrete, then the theoretical mutual information $I(X;T)=H(T)-H(T|X)$ is a constant since neural networks mapping samples bijectively. While if we assume the output feature space is continuous, then $I(X;T)$ would become positive infinity. Moreover, high dimensional data itself is unavoidably ill-posed and degenerated distributed, which complicates the definition of informative functionals and their estimation. These problems challenge many estimators, while some other approximation methods (i.e. Kernel-Density Estimation, Stochastic Binning) using additional noise to produce Stochastic Neural Networks, or dividing samples into discrete groups to avoid bijection (i.e. Binning, K-nearest). Unfortunately, Most methods revealed limitations such as sensitive to dimensions, samples variance and hyper-parameters\cite{geiger2020information}\cite{schiemer2019revisiting}\cite{cheng2019utilizing}.

Entropy estimation is a classical problem of information theory. Prior works have raised many estimators with LogDet function, such as uniformly minimum variance unbiased (UMVU)\cite{ahmed1989entropy}, bayesian method\cite{misra2005estimation}\cite{srivastava2008bayesian} and more recently nonparanormal estimator \cite{singh2017nonparanormal}. However, these methods often involve complex optimization, which is unsuitable for synchronous estimation in machine learning. Recent proposed matrix-based estimator $\alpha$-$R\acute{e}nyi$ entropy\cite{giraldo2014measures}\cite{yu2019multivariate} achieve good performance in deep neural networks\cite{yu2019understanding}\cite{yu2020understanding}\cite{yu2021measuring}\cite{tapia2020information}. However, its still lack of verification when applied in different models. In this work, we fully verified our proposed method. The proposed LogDet estimator measuring entropy with continuous samples with added noise. Such a continuity hypothesis enables us to directly approximate differential entropy.                                   Also, adding noise avoids the singular covariance matrix in LogDet estimator, which commonly appear when with finite samples or strongly correlated features. Further, we show LogDet estimator's performance by estimating entropy and mutual information. Then compare it with other commonly used methods. Results demonstrate that it is accurate, noise-robust and applicable to any dimension of features space, without worrying about saturation.

Apart from the entropy estimation problem, current analytical works report inconsistent conclusion about IB compression. Where its existence is observed in \cite{cheng2019utilizing}\cite{chelombiev2019adaptive} and absence is also argued in \cite{gabrie2019entropy}\cite{ goldfeld2018estimating}\cite{ li2021information}. Moreover, the promised connection of compression and generalization is also unclear. These results are recently accused corrupted by misused estimators\cite{schiemer2019revisiting}\cite{geiger2020information}, and other works argue that some meaningful results are connected with geometrical clustering\cite{tapia2020information}\cite{goldfeld2018estimating}\cite{schiemer2019revisiting}, rather than mutual information behaviour. Most of all, these debates stuck in arguing the existence of compression, but lack of sufficient verification of estimators. This prohibits the soundly investigation of compression. Some works try to provide theoretical analysis\cite{hafez2019compressed}, however, the area still lacks of empirical understanding. Also in need is the interpretation of compression, which is still unclear within current literatures.

After verifying the LogDet estimator, we believe one difficulty that prevents the understanding of IB behaviour comes from the complexity of $I(X;T)$ itself. Because even with a correct estimation of multi-layer neural networks, directly interpreting $I(X;T_i)$ of layer $i$ involve considering all former layers. Then the deeper $T_i$ we like to discuss, the more complex its behaviour will be. Taking the second layer as an example, in this case, whatever we observe in $I(X;T_2)$ is, it is difficult to confirm whether it is caused by the second or the first layer. Thus, we propose the concept of Layer Transmission Capacity (LTC) which estimates neural networks' capacity to transmit information in a layer-wise fashion. Based on this, we can discuss mutual information within a single layer. Our results give a direct interpretation of observed compression and provide a solid basis to its understanding.

Our main contributions can be concluded as bellow:
\begin{itemize}
\item[-] We propose the LogDet entropy estimator, which resolve the inaccurate analysis of misused estimators in previous works, and proved can be applied in any high dimensional feature spaces of deep neural networks.
\item[-] We compare different entropy estimators, which shows LogDet is robust to noise, large variance of samples and saturation problems.
\item[-] We proposed Layer Transmission Capacity (LTC) to provide a more precise mutual information of each layer in IB theory. It reveals IB compression is mostly an effect of the first layer, which has distinct behaviour against other layers in neural networks.
\end{itemize}

\section{Method}
Differential Entropy is the natural extension of discrete Shannon Entropy $H(X)=-\sum_{X}p(x)\log{p(x)}$. By replacing probability function with its density and integrating over x, the definition of differential entropy is: $h(X)=-\int_Xf(x)\log{f(x)}\rm{dx}$. Estimating differential entropy has long been an active area in Information Theory \cite{gupta2010parametric}\cite{poczos2012nonparametric1}\cite{srivastava2008bayesian}\cite{ ahmed1989entropy}. In principle, the differential entropy of multivariate Gaussian distribution can be represented by the logarithm determinant of its covariance matrix. Here we first introduce a approximation method LogDet for estimating differential entropy and define basic information measurements in multivariate Gaussian distribution. Next, we extend this method in multi
-signal situation and proposed the layer-wise estimation Layer Transmission Capacity of neural networks. Finally, we interpret LogDet estimator from the perspective of Coding Length Function\cite{ma2007segmentation}, and discuss the parameter setting.

\subsection{Approximating Differential Entropy with LogDet estimator}
The Logarithm Determinant (LogDet for short) is considered strongly connected with information entropy\cite{cover1999elements}. To a multi-variable Gaussian Distribution $X\sim{N}(\mu,\Sigma_X)$, the Shannon differential entropy can be represented by the log determinant of its covariance matrix as:
\begin{equation}
\begin{aligned}
     \mathbf{h}(X)&=\frac{1}{2}\log\det\Sigma_X+\frac{d}{2}(\log2\pi+1)\\
    &=\frac{1}{2}\sum_i^d\log\lambda_i+\frac{d}{2}(\log2\pi+1)
    \label{eq1}
\end{aligned}
\end{equation}
where $d$ is the dimension of variables and $\lambda_i$ denote the $ith$ eigenvalue of its covariance matrix. Obviously the second term is constant and the measurement is determined by $\log\det{\Sigma_x}$. Later on, by extending the generalization of Strong Sub-additive (SSA) condition\cite{adesso2016strong}, it reveals that all matrix monotone functions $f$ would satisfy the generalized SSA inequality: $Trf(V_{AC})+Trf(V_{BC})-Trf(V_{ABC})-Trf(V_{C})\geq{0}$. When having $\log{x}$ as $f(x)$, differential $R\acute{e}nyi$'s $\alpha$-order entropy \cite{renyi1961measures} of multivariate Gaussian is also equivalent to the $\log\det$ representation\cite{lami2017log}, which is:
\begin{equation}
\begin{aligned}
     \mathbf{h}_\alpha(X)&=\frac{1}{1-\alpha}\log{\int d^nxp_A(x)^\alpha}\\ &=\frac{1}{2}\log\det\Sigma_X+\frac{n}{2}(\log2\pi+\frac{1}{\alpha-1}\log{\alpha}).
    \label{eq2}
\end{aligned}
\end{equation}
Since $\alpha$-$R\acute{e}nyi$ entropy is the generalization of a family of entropy measurements, the above expression reveals the potential of LogDet function as a universal entropy measurement to multivariate random Gaussian variables. Indeed, $\log\det$ function has been acting as an ideal matrix-based estimator for decades, applications such as Jensen-Bregman LogDet Divergence \cite{cherian2012jensen}, Alpha–Beta and Gamma Divergences\cite{cichocki2015log} show the good properties to semi-definite matrices as robust to noises, positional bias and outlines, while maintaining the computational efficiency.

Despite the advantages, LogDet differential entropy is faced with two challenges: \textbf{(1)} First, when having finite but strongly correlated variables, the covariance matrix is often singular which leads to zero determinant. Similar cases often appear in current machine learning, where samples tend to be sparse and strongly correlated. \textbf{(2)} Denoting the continuous model's input signal as $X$ and intermediate variable as $T$, then $T$ is continuously determined by $X$. When $X$ is settled, the conditional probability density of each determined variable $T$ becomes a delta function. Therefore, the conditional entropy $H(T|X)$ would be negative infinite and lead to a positive infinite $I(X;T)=H(T)-H(T|X)$\cite{amjad2019learning}\cite{goldfeld2018estimating}\cite{saxe2019information}. This proves that estimating differential entropy directly in neural networks only returns the bias of estimators, rather than meaningful information entropy.

Prior works of entropy estimators provide a solution. By adding independent Gaussian noise $z\sim{N(0,I)}$ to each deterministic intermediate state $T$ as $\hat{T}=T+z$, we have a stochastic $\hat{T}$ to any given $X$. In such cases, deterministic problems is avoided and we can apply $\log\det$ operation to covariance matrix $Cov(\hat{T})=\Sigma+I$ to estimate its entropy. To eliminate artificial bias brings by added noise, rather than scale down noise's variance (which will cause a negative estimated entropy), we enlarge the original covariance matrix by an expanding factor $\beta$ to decrease the comparable weight of added noise. This leads to the following definition:

$Definition$ Given $n$ samples $X=[x_1,x_2,...,x_n]$ where each sample have $d$ dimensional features that $X\in{\mathbb{R}^{n\times{d}}}$, its covariance matrix is calculated by $\Sigma=\frac{X^TX}{n}$, the LogDet entropy can be represented as:
\begin{equation}
    \mathbf{H}_D(X)=\frac{1}{2}\log\det(I+\beta\Sigma)=\frac{1}{2}\log\det(I+\beta\frac{X^TX}{n})
    \label{eq3}
\end{equation}
where $\beta$ is the scaling parameter. Here, we show such definition describe the differential entropy. By performing SVD to $X$, Eq.\ref{eq3} is equivalent to:
\begin{equation}
\begin{aligned}
    \mathbf{H}_D(X)&=\sum_{i=1}^{k}\frac{1}{2}\log(1+\beta{\lambda_i})\\
    &=\sum_{i=1}^{k}\frac{1}{2}\log(\frac{1}{\beta}+\lambda_i) + \frac{k}{2}\log\beta
\end{aligned}
\label{eq4}
\end{equation}
where $k=rank(X)$ and $\lambda_i$ denotes the $ith$ eigenvalue of $\frac{X^TX}{n}$. We see the first term of Eq.\ref{eq4} approximate the dominant term in Eq.\ref{eq1}, which is the differential entropy of multivariate Gaussian random variables (RVs). The second term is a constant $\frac{1}{2}\log\beta$ weighted by $k$, which also change simultaneously with differential entropy. Therefore, LogDet is capable to estimate entropy of multivariate Gaussian RVs by approximating the differential entropy. (Details can be found in Appendix. A).

Showing the approaching property of LogDet estimator, we can further define joint LogDet entropy as:

$Definition$ Given samples sets $X_1\in\mathbb{R}^{n\times{d_1}}$,$X_2\in\mathbb{R}^{n\times{d_2}}$, if denote $Z = [X_1,X_2]\in\mathbb{R}^{n\times(d_1+d_2)}$ as a concatenation of $X_1$ and $X_2$, and its covariance matrix is $\Sigma_{Z}=\begin{pmatrix} \Sigma_1 & F \\ F^T & \Sigma_2 \end{pmatrix}$, where $\Sigma_1$ and $\Sigma_2$ is covariance matrices of $X_1$ and $X_2$ respectively, and $F$ equals to $\frac{X_1^TX_2}{n}$, the joint LogDet entropy of $X_1$ and $X_2$ is:
\begin{equation}
\begin{aligned}
    \mathbf{H}_D(X_1,X_2)&=\frac{1}{2}\log\det(I+\beta\Sigma_{Z})\\
    &=\frac{1}{2}\log\det(I+\beta\frac{Z^TZ}{n})
    \label{eq5}
\end{aligned}
\end{equation}.

Some prior works summarize determinant inequalities from the aspect of information theory\cite{cover1988determinant}\cite{matic2014inequalities}\cite{cai2015law}. These inequalities gives a theoretical background to define other LogDet information measurements.

$Proposition$ Given random variables $X_1$ and $X_2$, their joint LogDet entropy satisfies the following inequalities (proof see Appendix. A):
\begin{align}
     \mathbf{H}_D(X_1,X_2)&\leq\mathbf{H}_D(X_1)+\mathbf{H}_D(X_2)\\
     \mathbf{H}_D(X_1,X_2)&\geq\max(\mathbf{H}_D(X_1)+\mathbf{H}_D(X_2))
     \label{eq6}
\end{align}

Therefore, we can further compute conditional entropy and mutual information as:
\begin{align}
    &\mathbf{H}_D(X_1|X_2)=\mathbf{H}_D(X_1,X_2) - \mathbf{H}_D(X_2)\\
    &\mathbf{I}_D(X_1;X_2)=\mathbf{H}_D(X_1)+\mathbf{H}_D(X_2)-\mathbf{H}_D(X_1,X_2)
    \label{eq7}
\end{align}.

\begin{figure}
\centering
\subfloat[$I_D(X;Y)$ via Dimensions with n=128]{
\includegraphics[width=0.5\textwidth]{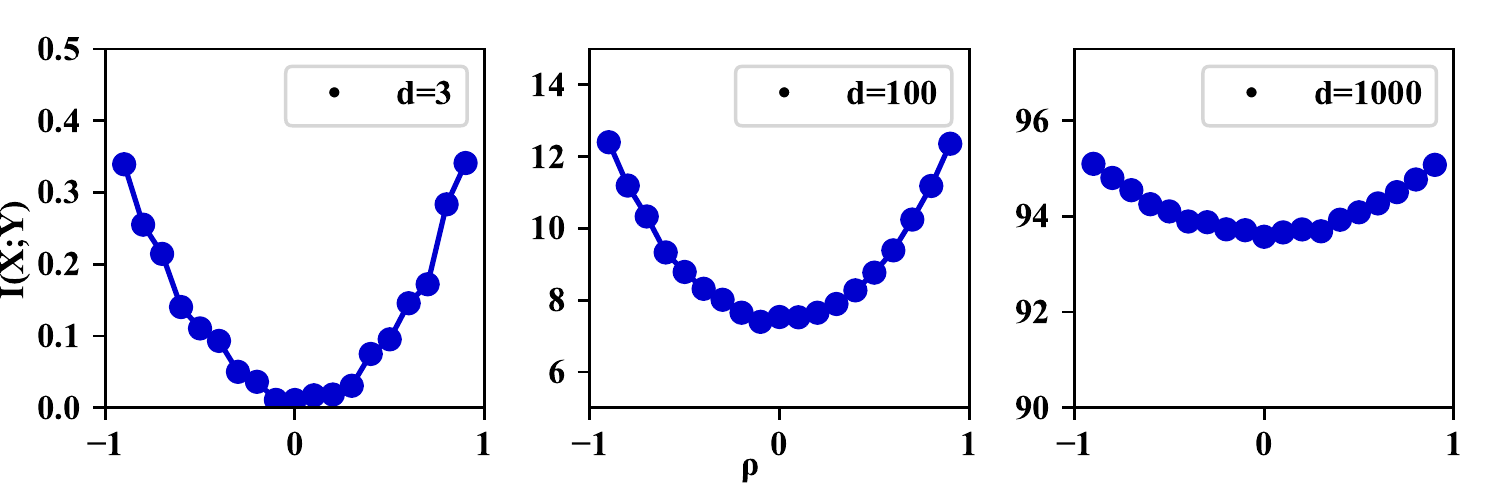}%
}
\hfil
\subfloat[$I_D(X;Y)$ via Sample Number with d=1000]{
\includegraphics[width=0.5\textwidth]{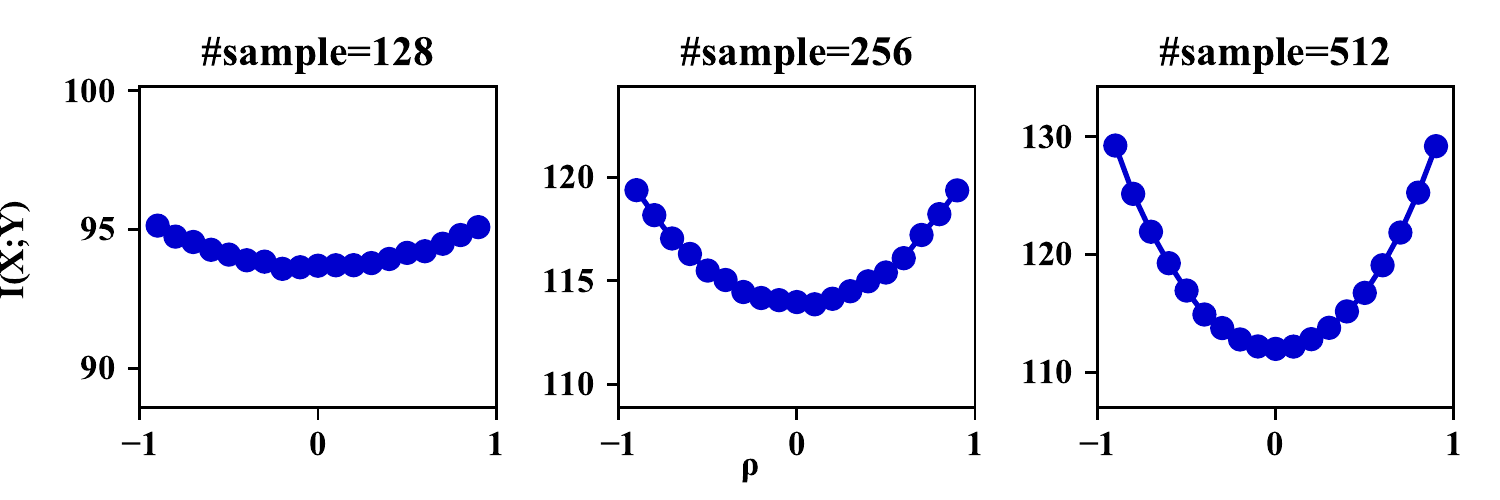}%
}
\caption{Experiment of $I_D(X;Y)$ with multivariate Gaussian samples. $d$ is feature dimensions, and correlation factor $\rho$ grows from -1 to 1. We set $\beta=1$ in this case. All curves track the theoretical tendency of mutual information.}
\label{MI}
\end{figure}
We test Mutual information $I_D(X_1;X_2)$ using a experiment in Fig. \ref{MI}. $X$ and $Y$ are $d$ dimensional multivariate RVs that obey $N(0,I)$, while variables between $X$ and $Y$ is controlled correlated with factor $\rho$ that $Cov(X,Y)=\rho{I}$. In other words, the covariance between RVs of $X$ and those of $Y$ is $\rho$. We alter $\rho$ from -1 to 1, and estimate the mutual information $I_D(X;Y)$ with LogDet estimator. Similar experiment is adopted in \cite{tapia2020information}\cite{belghazi2018mine}. Besides, we repeat this experiment on samples with different dimension $d\in$\{3,100,1000\}. Experimental results show that the estimated $I_D(X;Y)$ is not exactly theoretical mutual information, and the minimum value is also not zero when $\rho=0$. But clearly, all curves follow the tendency of theoretical mutual information, leading to a concave shape. The result also suggest LogDet estimator can be applied in high-dimensional space without saturation problems. In Fig. 1(b), we see adding more samples improves the result, whose curve shows deeper reduction than those with fewer samples. In all, we can use LogDet estimator to approximate mutual information.

% ------------------------------------ ------------------------------------
\subsection{Multi-signal Extension}
Beside the basic information measurements, LogDet estimator can be extended to other complex data structures. For example, for multiple signals $X=[X_1,X_2,...,X_k]$, where each $X_i$ is a multivariate RV. This condition is common neural networks, such as multi-channel convolutional neural networks, since features in different channel are often calculated by different parameters, therefore, instead of regarding the whole feature map as a multivariate RV, it is better to split apart features from different channels as multiple signals, and calculate their joint entropy. Here we will show LogDet estimator can be naturally extended to such multiple-signal conditions by multivariate joint entropy.

$Definition$ Given the variable set $Z=[X_1,X_2,X_3,...,X_k]$ where $X_i$ is a multivariate RV, denote the covariance matrix between $X_i$ and $X_j$ as $\Sigma_{i,j}$, we then have the covariance matrix of 
$Z$ equals to $\Sigma_{1,2,...,k}=\begin{pmatrix} 
\Sigma_{1,1}     & \Sigma_{1,2} & \cdots & \Sigma_{1,k}\\
\Sigma_{2,1} & \Sigma_{2,2}     &        &\vdots \\
\vdots       &              & \ddots \\
\Sigma_{k,1} & \cdots       &        &\Sigma_{k,k}
\end{pmatrix}$,
the joint entropy of $[X_1,X_2,...,X_k]$ is given by:
\begin{equation}
    \begin{aligned}
         \mathbf{H}_D(X_1,X_2,...,X_k) = \frac{1}{2}\log\det(I+\beta\Sigma_{1,2,...,k})
    \end{aligned}
\end{equation}
The definition above directly extend joint entropy with LogDet estiamtor. However, the size of covariance matrix of multiple signal $Cov(Z)=\Sigma_{1,2,...,k}\in\mathbb{R}^{kd\times{kd}}$ can rapidly increase by signal number $k$ and feature dimension $d$, which is expensive to perform $\log\det$ operation on and to save. Luckily, by applying $\log\det$'s commutative property\cite{ma2007segmentation}, it suggests an equivalent approach to calculate the multi-signal entropy. Having $n$ estimating samples for each variable $X_i$, the above expression equals to:
\begin{equation*}
    \begin{aligned}
    \mathbf{H}_D(X_1,X_2,...,X_k)&=\frac{1}{2}\log\det(I+\beta\frac{Z^{T}Z}{n})\\
    &=\frac{1}{2}\log\det(I+\beta\frac{ZZ^T}{n})\\
    &=\frac{1}{2}\log\det(I+\frac{\beta}{n}\sum_{i=1}^{k}X_{i}X_i^T)
    \end{aligned}
\end{equation*}
Under this expression, the estimated matrix size become independent to $X_i$'s feature dimension and signal number, limiting its size within $\mathbb{R}^{n\times{n}}$. Therefore, the complexity only grows linearly with signal numbers $k$. Moreover, this interpretation suggests covariance matrix in LogDet entropy is equivilent to the sum of $X_iX_i^T$. Which is the pair-wise similarity matrix that contains samples' inner-production as components. This means we can naturally introduce kernel methods to adjust LogDet estimator to more complex input distribution. After all, we extend the LogDet estimator to multi-signal situation and discover a efficient approach in application.

% Is the above proposition persuasive enough?

Under the above expression, the following properties can be derived with Minkowski determinant theorem and the quality of $\log\det$ function to positive semi-definite matrix(proof See Appendix. A). 

$Corollary$
Segmenting the variable set $X=[X_1,X_2,...,X_k]$ into arbitrary number of subsets, denoted as [$X^{(1)}, X^{(2)},...,X^{(t)}$], where each allocated $X^{(i)}$ is the complement to the concatenation of the rest as [$X^{(1)}, X^{(2)},...,X^{(i-1)},X^{(i+1)},...,X^{(t)}$] with respect to original variable set $X=[X_1,X_2,...,X_k]$, the following inequalities hold:
\begin{align}
    \mathbf{H}_D(X) &\leq \mathbf{H}_D(X^{(1)}) + \mathbf{H}_D(X^{(2)})+...+\mathbf{H}_D(X^{(t)})\\  
    \mathbf{H}_D(X) &\geq \max(\mathbf{H}_D(X^{(1)}), \mathbf{H}_D(X^{(2)}),...,\mathbf{H}_D(X^{(t)}))
    \label{Comv}
\end{align}
Where the first equality holds i.f.f. variables between each subset is fully independent that has zero covariance, which means $\Sigma_{i,j}=0$ for any $X_i$ and $X_j$ ($i\neq{j}$) that belong to different subsets. The second inequality holds i.f.f. all the rest variables are fully determined, providing no information to the largest one. We also notice that, when with equal $X_i$ and $X
_j$, the joint estimation only increase with a constant $k\log{2}$ where $k$ is the matrix rank. It can be interpreted as representing the same variables only requires additional $k$ bits for marking the equivalence of variables between $X_i$ and $X_j$. These extended inequalities and equality conditions are fully aligned with the definition of information entropy, therefore, providing a theoretical guarantee for our estimation.

% With the Joint Entropy of multiple variables, we can construct a wide range of information estimation. Three variables for example, we can defined the conditional mutual information as:
% \begin{equation}
% \begin{aligned}
%     \mathbf{I}_D(X_i,X_j|X_k) = \mathbf{H}_D(X_i,X_k) + \mathbf{H}_D(X_j,X_k) \\- \mathbf{H}_D(X_i,X_j,X_k) - \mathbf{H}_D(X_k)
% \end{aligned}
% \end{equation}

\begin{figure*}
\centering
\includegraphics[width=0.8\textwidth]{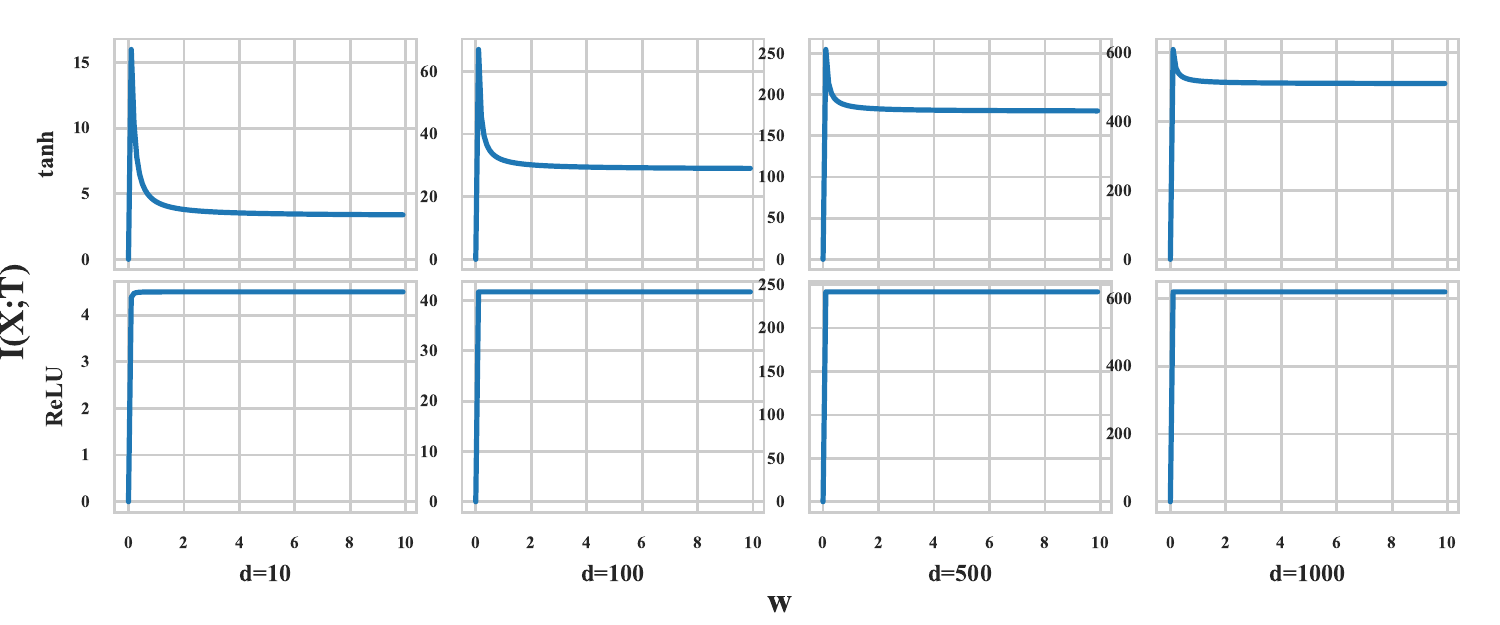}
\caption{Estimating activation function's saturating effect using LogDet mutual information estimator. Input $X$ is weighted by $w$ and operated by $tanh$ and $ReLU$ respectively. More value would saturated in $tanh$ when $w$ grows, revealed as decrease in $I(X;T)$. The experiment is conducted in different dimension $d$, and using 3000 samples.}
\label{actcomp}
\end{figure*}

\subsection{Layer Transmission Capacity} 
The computing process of neural networks from input to intermediate representation $X\xrightarrow[]{}T$ often involves operations of numerous layers, it is hard to locate which layer is responsible for changes in $I(X;T)$. Therefore, we propose an estimation of Layer Transmission Capacity to simplify the discussion of mutual information behaviour. By measuring mutual information of each layer's input and output, we can quantify how much information is transmitted through. This enables the discussion of each layer's behaviour without most of interference, here we give its intuitive understanding from the perspective of Gaussian Channels.

To any neural network, each layer has two main components: (1) Parameters $\theta$ (2) Activation function $g(.)$. During computations, Input $T$ is weighted by the parameters $\theta$, then activated by $g(.)$ as $L(T)=g(\theta\cdot{T})$. Thereafter, $\theta$ is updated according to the gradient feedback. Assuming we have an infinite number of input $T$ that depend on a continuous distribution of task data, then each layer can be modeled with Gaussian Feedback Channel\cite{cover1999elements}, where model parameters $\theta$ function as an explicit encoder, and the activation function as special 'narrow' channel that has structured interruption to encoded words $\theta\cdot{T}$. Such interruption solely depends on the non-linearity of each activation function and can therefore considered as coloured noise, which is not independent Gaussian but with a particular Phase Structure. Naturally, the back-propagated gradients are feedback to the encoder $\theta$, guiding the codebook design.

Under the above description, assuming we have a continuous input signal $T$, layer $L$'s simulated Channel Capacity is defined as the largest amount of information transmitted through. which is:
\begin{equation}
\begin{aligned}
    C &= \max_{p(T):E(T^2)\leq{P}} I(T; L(T))\\
\end{aligned}
\end{equation}
where $L(T)$ is the output of layer $L$, P is variance constrain of channel input $T$. This tells that, with proper distribution $p(T)$, we can precisely calculate how much information layer $L$ can transmit as most. Since we only need to consider $T$ from the task distribution, and the explicit encoder $\theta$ can be optimized through training. Therefore, the Capacity of $L$ should be maximized concerning $\theta$ and $p(T)$. When the task distribution $p(T)$ is settled (by choosing the training data in DNNs), we can track the dynamic of such transmitted capacity as a function to parameter $\theta$. We denote this function as Layer Transmission Capacity (LTC):

$Definition$
Given sufficient observing samples $X$, to a neural network layer $L$, whose input denoted as $T$ and output as $L(T)$, the Layer Transmission Capacity of $L$ is given by:
\begin{equation}
    LTC(L)[\theta] = I_D(T; L(T;\theta))
\end{equation}

It is worth notice that, LTC cannot fully remove the influence of the former layer, because $I_D(T;L(T))$ can still be affected by $H_D(T)$, which is determined by the former layer. However, it remove most elements that irrelevant to layers' operation and concentrate the discussion in the current layer, which achieve us a better understanding. Also when the task is settled, each layer's input $T$ will have a converged distribution, so function $LTC(L)[\theta]$ have an upper bond when maximized with $\theta$. This ensures the value is varied within a constraint range, rather than change infinitely. After all, we can use Layer Transmission Capacity function to evaluate the transmitting behaviour of each layer.
% bn act compare

\subsection{Relation to Coding Length Function}

% \begin{figure}
% \centering
% \includegraphics[height=6.6cm,width=11.2cm]{Img/Figure1.pdf}
% \caption{The Visualization of Coding Length Function from \cite{ma2007segmentation}. By dividing the whole sample space $vol(\hat{W})$ with the tolerable error range $vol(\epsilon^2I)$, and label each one with binary code, then each sample will be allocated with the position code with logarithm of the total unit number. Therefore, we can identify any sample's position with its allocated binary code. Reduce $\epsilon$ indicate smaller error range, at a cost of increasing coding number.}
% \label{CLF}
% \end{figure}
Differential entropy gives the amount of binary bits to encode continuous variables, however in practice, it is difficult to represent arbitrary continuous RVs with finite codes. Coding Length Function provides a substitution\cite{ma2007segmentation}, which measures the distortion rate from Rate Distortion Theory. Our approach can also be considered as an extension of the Coding Length Function. Here we briefly review the Coding Length Function, and interpret our LogDet estimator to show how $\beta$ is related to reducing distortion ratio. This discussion also helps to select a proper $\beta$.

\paragraph{Coding Length Function}
Coding length function is raised to estimate Distortion Rate from Rate Distortion Theory\cite{cover1999elements}. Despite the difficulty to encode continuous RVs into finite representation, it is still proved approachable by allowing a small error as encoding distortion. Then, the distortion rate function gives the minimum number of bits (or distortion rate) of such codes. Formally, the distortion rate is defined as:
$$R(D)=R^{(I)}(D)=\min_{f:d(\mathbf{x},\mathbf{\hat{x}})=D} \mathbf{I}(\mathbf{x};\mathbf{\hat{x}})$$
where $\mathbf{x}$ is the original signal, and $\mathbf{\hat{x}}$ is its distorted representation, with distortion $D$.
Then, the Coding Length Function of $n$ samples from arbitrary distribution $\mathbf{W}=[\mathbf{w}_1,\mathbf{w}_2,...,\mathbf{w}_n]$ where $\mathbf{W}\in \mathbb{R}^{n\times{d}}$ is given by:
\begin{equation}
    \begin{aligned}
         R(W)=\log_2(\frac{vol(\hat{W})}{vol(z)})=\frac{1}{2}\log\det(I+\frac{d}{n\epsilon^2}W^TW)
    \end{aligned}
    \label{eq9}
\end{equation}
$\epsilon$ is the allowable distortion. This formula is achieved by dividing the volume of distorted samples $vol(\hat{W})=\sqrt{\det(\frac{\epsilon^2}{d}I+\frac{W^TW}{n})}$ with the noise sphere $vol(z)=\sqrt{\det(\frac{\epsilon^2}{d})}$, and labeling the divided spheres with binary codes. Therefore, each sphere can be localized with their position code, and represent samples within the distortion radius. The length of such code gives the smallest number of bits to represent the distorted sample.

\begin{figure*}
\centering
\subfloat[$H_D(X)$ Saturation Test]{
\includegraphics[width=0.45\textwidth]{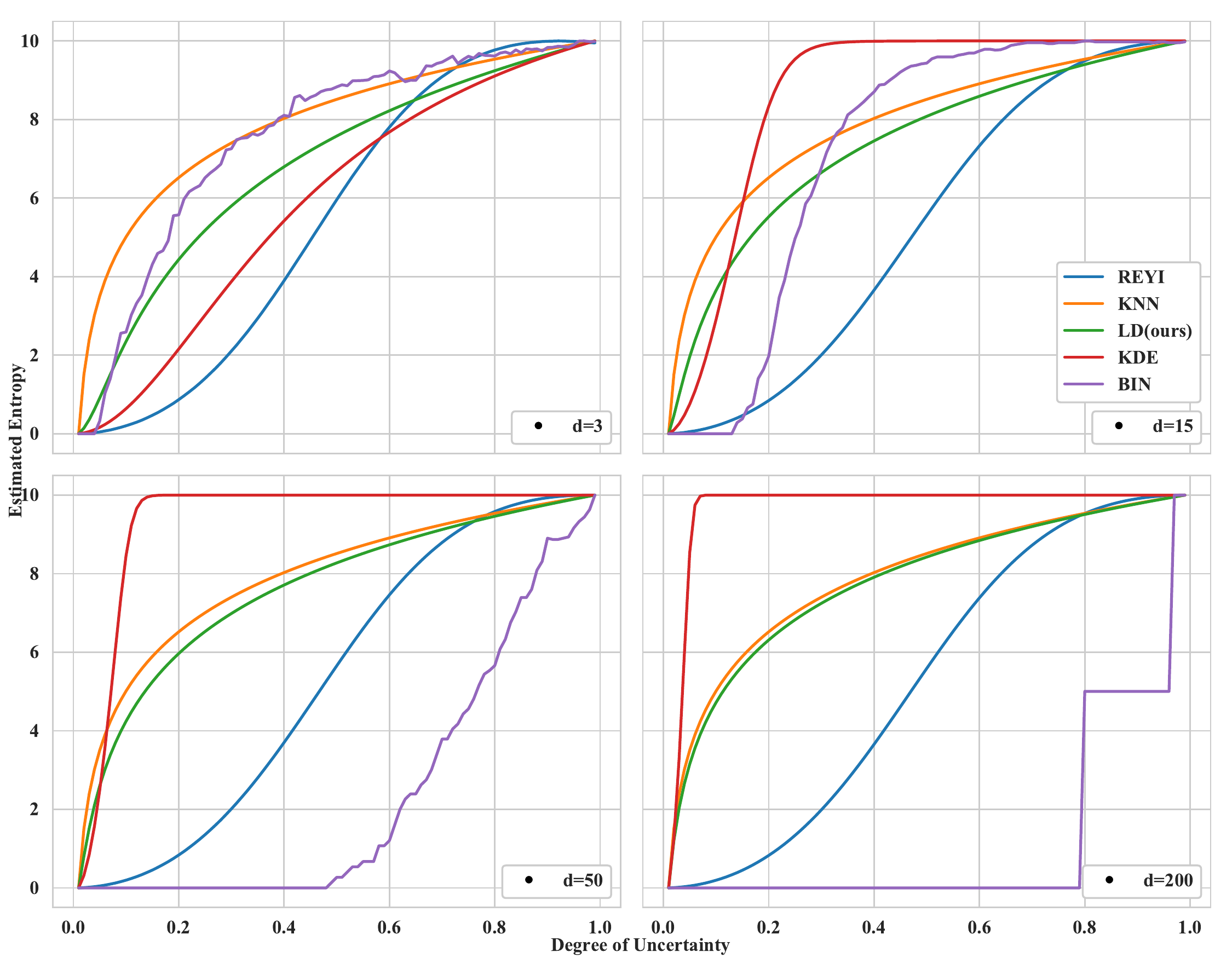}
}
\subfloat[$H_D(X)$ Precision Test]{
\includegraphics[width=0.45\textwidth]{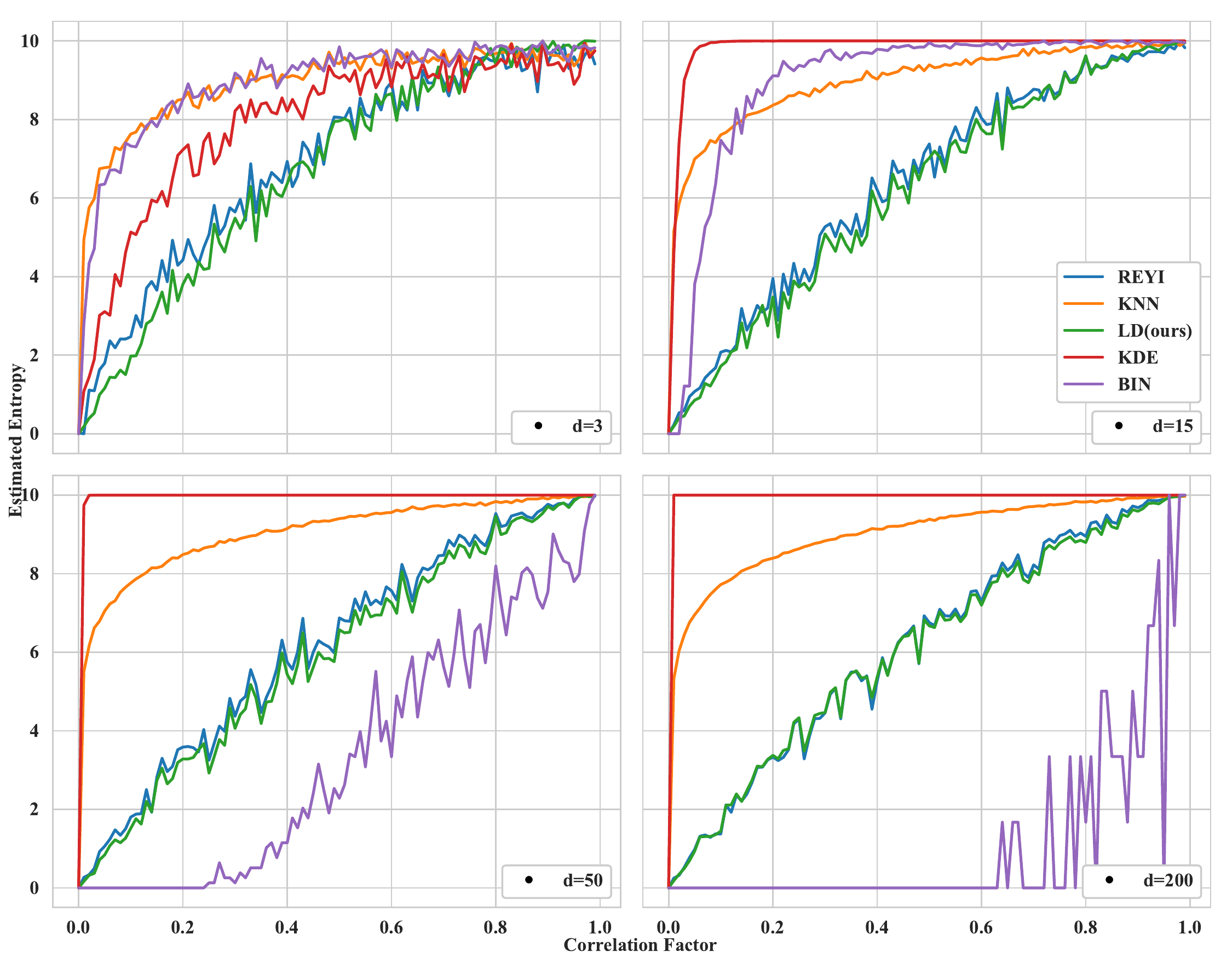}
}
\caption{Estimator's saturation test (a) and precision test (b) comparing with different estimation methods (REYI\cite{yu2019multivariate,tapia2020information}, KNN\cite{kraskov2004estimating,ver2013information}, KDE\cite{saxe2019information}, BIN\cite{cheng2018evaluating} and ours LD(LogDet)). In Saturation Test, samples are varied with increasing variance (from 0 to 1). In Precision test, samples $X$'s variance is fixed in 1, the curve is plotted along the inner correlation factors $\rho$, where $Cov(x_i,x_j)=\rho$ for $i\neq{j}.$ For comparison, all estimated result is unified in range of (0,10). 
In (a), BIN, KDE saturating when increased the dimension. And in (b), only REYI, KNN and our method can precisely distinguish with correlated variables in all dimensional cases, but comparably, REYI and our method is observed better preciser increase than KNN. In all, LogDet can theoretically estimate variate samples with high precision in all dimensional cases.}
\label{dimension compare}
\end{figure*}

It is easy to find the consensus between Eq.\ref{eq9} and Eq.\ref{eq3}. Let $\beta=\frac{n}{\epsilon^2}$, then increasing $\beta$ to eliminate bias of LogDet entropy is equivalent to reducing distortion $\epsilon$ in $R(D)$. Since $R(D)$ as a measure of distortion rate, should approach the Shannon Entropy $\mathbf{H}$ when taking zero as limit of $\epsilon$. This proves the plausibility of LogDet's approximation method, explaining why an expand factor $\beta$ is essential for accurate measuring. Moreover, prior analysis of coding length function reveals it that, as a tight upper bond of coding length of arbitrary distributed variables, it can be applied in a wide range of machine learning tasks and datasets\cite{ma2007segmentation}\cite{chan2020deep}\cite{yu2020learning}\cite{wu2020incremental}. This shows LogDet's potential as a entropy measurement in realworld benchmarks.

 The Distortion Rate for multivariate Gaussian $R(D)=\log\det\frac{\Sigma}{D}$ requires that D smaller than any eigenvalue of covariance matrix $\Sigma$\cite{cover1999elements}\cite{ma2007segmentation}. This indicates a small $D$ as well as a large $\beta$ in LogDet estimator. In this work, we follow the prior applications in Coding Length Function\cite{yu2020learning}\cite{yu2021deep} where $D=\frac{\epsilon^2}{d}$, which successfully measures the compactness of samples. Therefore, we let $\epsilon=0.1$ so $\beta=100d$ in LogDet entropy $H_D(X)$. In $I_D(X_1;X_2)$, $\beta$ is set as $100(d_1+d_2)$, where $d_1$, $d_2$ are feature dimensions of $X_1$, $X_2$ respectively. 
 
 We adopt the test of activation function in \cite{saxe2019information} to verify whether LogDet under this setting can be used to estimate mutual information in neural networks. Since large variables value is saturated after activated by doule-saturating function $tanh$, the model output's connection with it of input is reduced. Such reduction of connection can be revealed as a decrease of $I(X;T)$ in neural networks. In this test, we enlarge the weight of a single fully connected layer, and expect LogDet estimator can track such decreases, which validates its capacity to estimate informative behaviour in neural networks. In Fig.\ref{actcomp}, such decreases occur in all tested feature dimension with $tanh$, and absents when with $ReLU$ activation. This is consistent with our expectation. It also shows that $LogDet$ mutual information estimator would not be affected by the sample variance, but solely reveal the change in variable dependency. In which case, we don't need to concern the disruption of variance changes during training. However, the result suggests a problem in estimating high dimensional samples. As a method relies on matrix estimation, the accuracy of $LogDet$ estimator and other matrix-based methods heavily influenced by number of estimated sample. In this paper's experiments, this test shows 3000 samples is sufficient. As for the limitation of matrix-based methods, we leave the discussion in Appendix. B.

\section{Experiments}

Experiments are divided into three parts. 
First, we use generated Gaussian samples to investigate LogDet's property when estimating high dimensional samples. Second, to show our method is applicable in real-world datasets, we design an experiment with Machine Learning benchmarks MNIST, CIFAR10 and STL10. Then, we utilize LogDet estimator for Information Bottleneck analysis in neural networks, and compare the result with Layer Transmission Capacity to give an empirical interpretation of compression phenomenon and Networks' information behaviour. In all estimation, each variable is shifted with mean zero.

The first two experiments also involve other commonly adopted estimators. Methods we compare include: the improved version\cite{tapia2020information} of recent proposed multivariate matrix based $R\acute{e}nyi$'s $\alpha$-order entropy(REYI) \cite{yu2019multivariate}; the high dimensional adaptation\cite{ver2013information} of classical K-nearest neighbour(KNN) method\cite{kraskov2004estimating}; Kernel Density Estimation(KDE) method\cite{saxe2019information} and the Binning(BIN) method\cite{cheng2018evaluating}.

\begin{figure*}
\centering
\includegraphics[height=15cm,width=17cm]{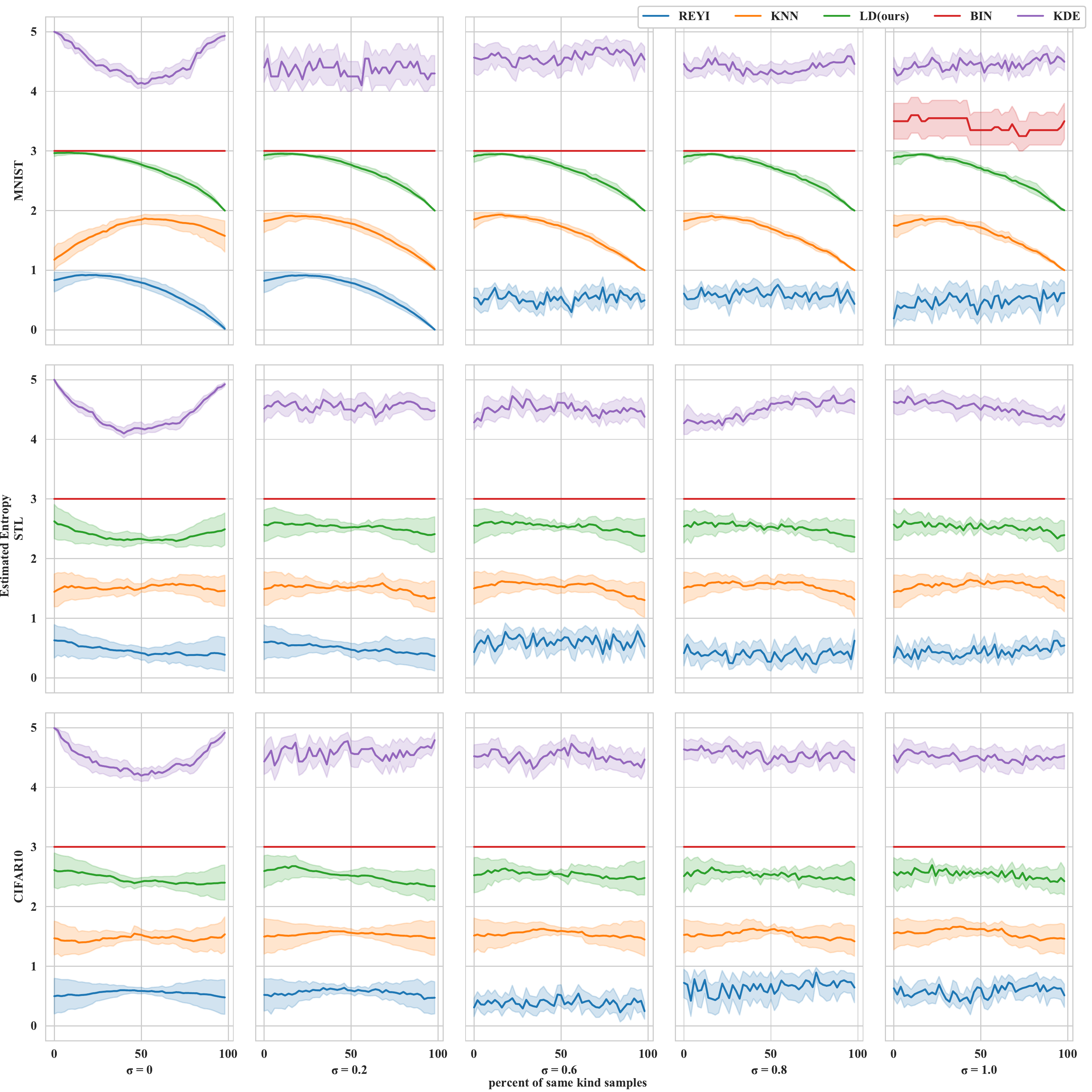}
\caption{Entroph Estimation using Machine Learning Image Benchmarks. Each row corresponds to one of three test datasets: MNIST, STL and CIFAR10, the columns is divided by the variance ratio $\sigma$ of added noise, from 0 to 1. All the value is unified between 0 and 1, and shifted in different vertical positions. The estimated value should decline as more single kind of sample appeared in the sample set.}
\label{degenerated}
\end{figure*}

\subsection{LogDet as Entropy Estimator}
Recall the fact that the definition of other informative functionals are based on LogDet entropy $H_D(X)$. Therefore, it is essential to validate whether LogDet can measure information entropy in different situation. Consider a multivariate random variable $X\sim{N(0,\Sigma)}$, its differential entropy $H(X)=\frac{1}{2}\log\det\Sigma+\frac{d}{2}(\log2\pi+1)$ is controled by two main components (determine diagonal and non-diagonal value of $\Sigma$ respectively): Variance and variables' Correlation. Therefore, we verify LogDet estimator's performance with different variances of independent Gaussian and changing Correlations of fix-variance samples. 
Moreover, this experiment gives a comparable test with other methods. Although each estimator often defines Mutual Information with different approximation methods (Such as the hypothesis of $I(X;T)=H(T)$, and KL-divergence definition of MI), the accurate entropy measure is still a common requirement of most estimators, in which case, by comparing their performance in this test, different estimators can be equally verified.

\begin{figure*}
\centering
\subfloat[IP of the FCN trained on CIFAR10]{
\includegraphics[width=0.45\textwidth]{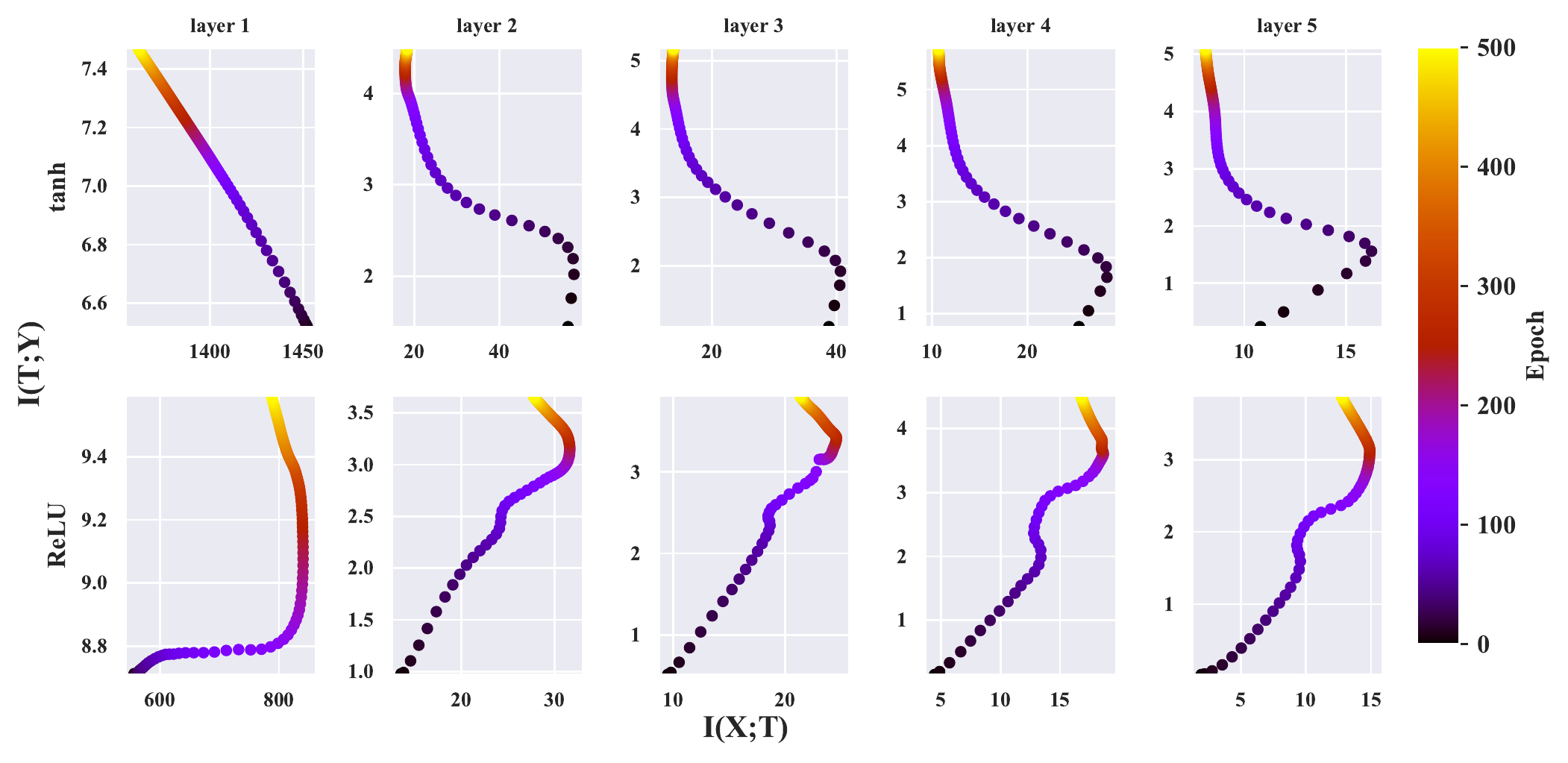}
}
\subfloat[IP of the FCN trained on MNIST]{
\includegraphics[width=0.45\textwidth]{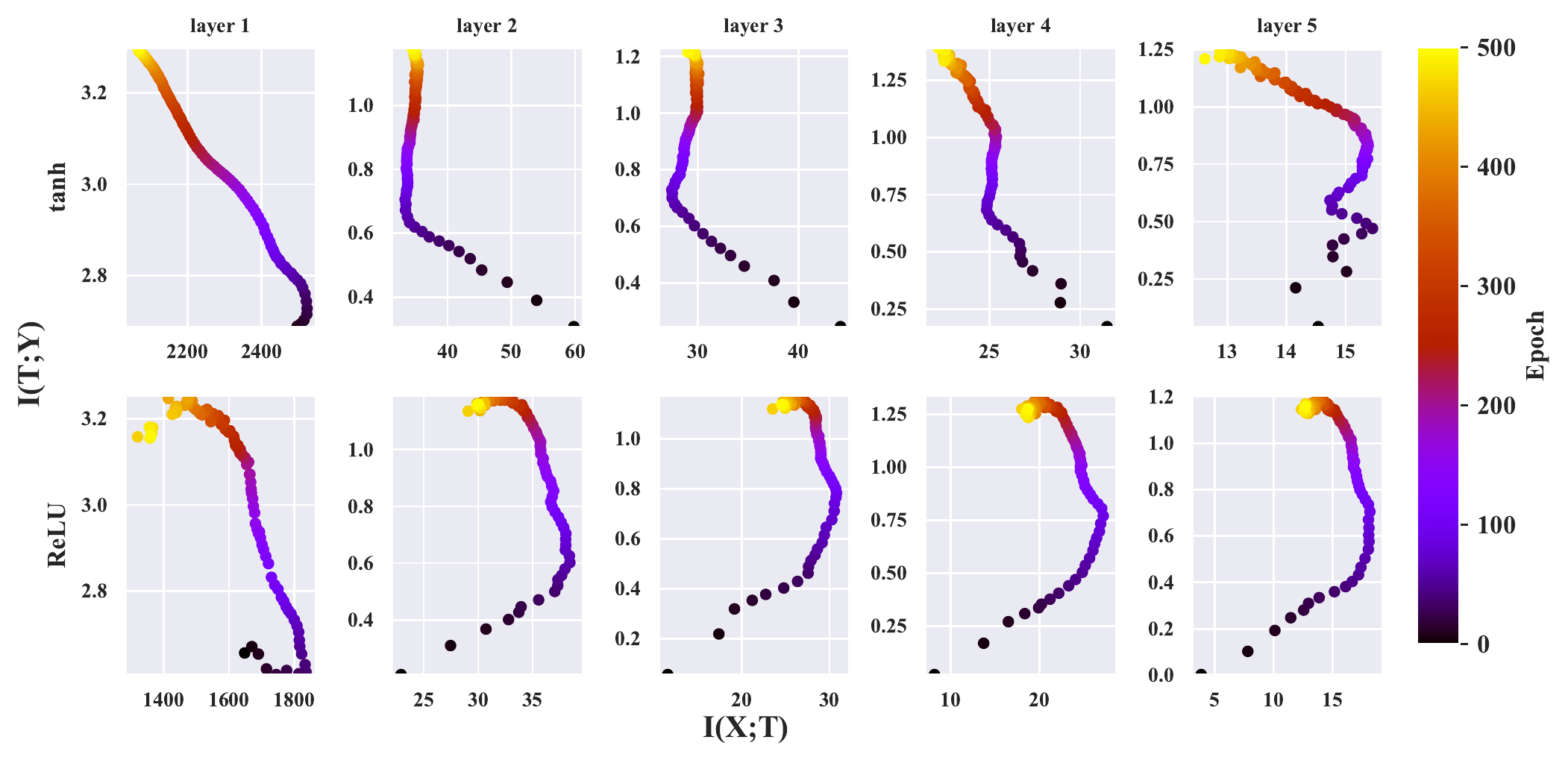}
}
\caption{Information Plane of a fully connected neural network with (784/3072-1024-20-20-10) neurons in each layer, trained on MNIST and CIFAR10 dataset. We sampled 100 from 500 epoch and displayed each layer separately.}
\label{fnn_rep}
\end{figure*}

We design two experiments with generated Gaussian samples. The first one maintains the independence of variables as $X\sim{N(0,\sigma{I})}$. We gradually increase $\sigma$ from 0 to 1, and show the curve of all compared method. By conducting this experiment in different feature dimension, we can verify whether these estimators suffer saturation problem with high variance samples. We denote this as Saturation Test in Fig.\ref{dimension compare}(a).

In the other test, samples are collected from a correlated multivariate Normal Distribution(i.e. the variance of each variable is controlled as one). Let $X\sim{N(0,\Sigma)}$, with $\Sigma=[s_{ij}]_{i,j=0}^d$ where $s_{ij}=1$ if $i=j$ and otherwise $s_{ij}=\sigma$. We decrease the correlation factor $\sigma$ from initial 1 to 0, as a change of correlation from fully correlated, to completely independent. The reduction of correlation should increase the estimated entropy. This experiment can find out whether estimators can reveal precise dependence change, can be regarded as a test of estimators' accuracy in Fig.\ref{dimension compare} (b). We also repeat the experiment in feature dimensions of [3,15,50,200].

In Fig.\ref{dimension compare} (a), it is clear that all methods can measure entropy with 3 variables. While in high dimensional cases, KDE(red) and BIN(purple) estimators experience saturation in different degree. KDE(red) method grows sharply with a slight increase of variance, indicating it is unable to distinguish samples with more diverse distributions. BIN(purple) method saturate in another way that can only distinguish those samples with high variances, revealing it would be inaccurate in most high dimensional estimation (because other samples are less diverse than Gaussian). In Fig.(b), all methods perform well in 3-dimensional case, but REYI(blue) and our LogDet(green) method have good precision when dimension increases. KNN(orange) method can also show the correlation change, but it grows rapidly in begins, which is less 
credible than REYI(blue) and LogDet(green).

Experiments reveal LogDet estimator can measure entropy in all dimensional feature space, without concern of saturation, and is accurate to measure the correlation between variables.

\subsection{LogDet Estimator on Realworld Benchmarks}

When measuring samples with high feature dimension, their distribution is unavoidably ill-posed and degenerated. Such as image data, for the strong correlation between pixels, images often contain vast redundancies which make them low-rank in their feature space. Meanwhile, the difference between images is hard to evaluate. For pictures in the same class, their quantified similarity often seems neglectable comparing to their differences. This makes conventional measurement hard to define on such data. However, estimating in such highly diversified and degenerated situation is a common requirement of modern machine learning applications. In Fig.\ref{dimension compare}, Our experiments show LogDet estimator can precisely measure entropy with different dimensional features, it is natural to wonder whether it can be applied in real world to reveal correlation in such degenerated distribution.

To find out, we choose 1000 samples from random classes in Machine Learning benchmarks MNIST (d=784), CIFAR10 (d=3072) and 500 samples from STL10 (d=3072). By gradually replacing the mix-class samples with images from a single category, we would expect the estimated entropy to decrease continuously. Because samples from the same class are more alike, their features would be more inner correlated, which indicates a lower entropy than their mix. We repeatedly perform the entropy estimation on each substitute class in each dataset. So we will have 10 results per dataset per method. By averaging the results, the bias brings by the entropy difference of each class can be eliminated. To compare each method's performance, we unify all values from 0 to 1 without changing their distribution. We also compare the result on the original image, and those with Gaussian noise by variance $\epsilon\in$[0.2, 0.6, 0.8, 1.0] to further analyse these estimator's robustness to random noise.

The experiment result is displayed in Fig.\ref{degenerated}. When tested with MNIST and STL10, LogDet estimator, $\alpha$-$R\acute{e}nyi$ estimator and K-nearest approach show expected reduction. However, when with added noises, $\alpha$-$R\acute{e}nyi$ is comparably more unstable. Its estimation fails when noise variance reached 0.6. While in CIFAR10, the decline of $\alpha$-$R\acute{e}nyi$ estimator and K-nearest approach become less obvious than LogDet. Despite the fluctuation, the general curve of LogDet estimator's is still downward. Besides these three methods, the Binning approach saturates in most cases. KDE fails in another way, showing a concave curve in all three unbiased tests. Then it is randomly fluctuated with added noises, suggesting it unsuitable to measure in image datasets.

Comparing the result, LogDet estimator shows statistically declines within all noise situation and all tested datasets. This shows its capability in estimating machine learning benchmarks with high-dimensional degenerated distribution, and is robust to random noise. We think the stability is brought by the logarithm determinant operation, which aligns with the findings of prior works about LogDet based matrix divergence \cite{cherian2012jensen}\cite{cichocki2015log}.

\subsection{LogDet Estimator for Information Behaviour Analysis}

\begin{figure*}
\centering
\subfloat[IP of the 3-layer FCN]{
\includegraphics[width=0.36\textwidth]{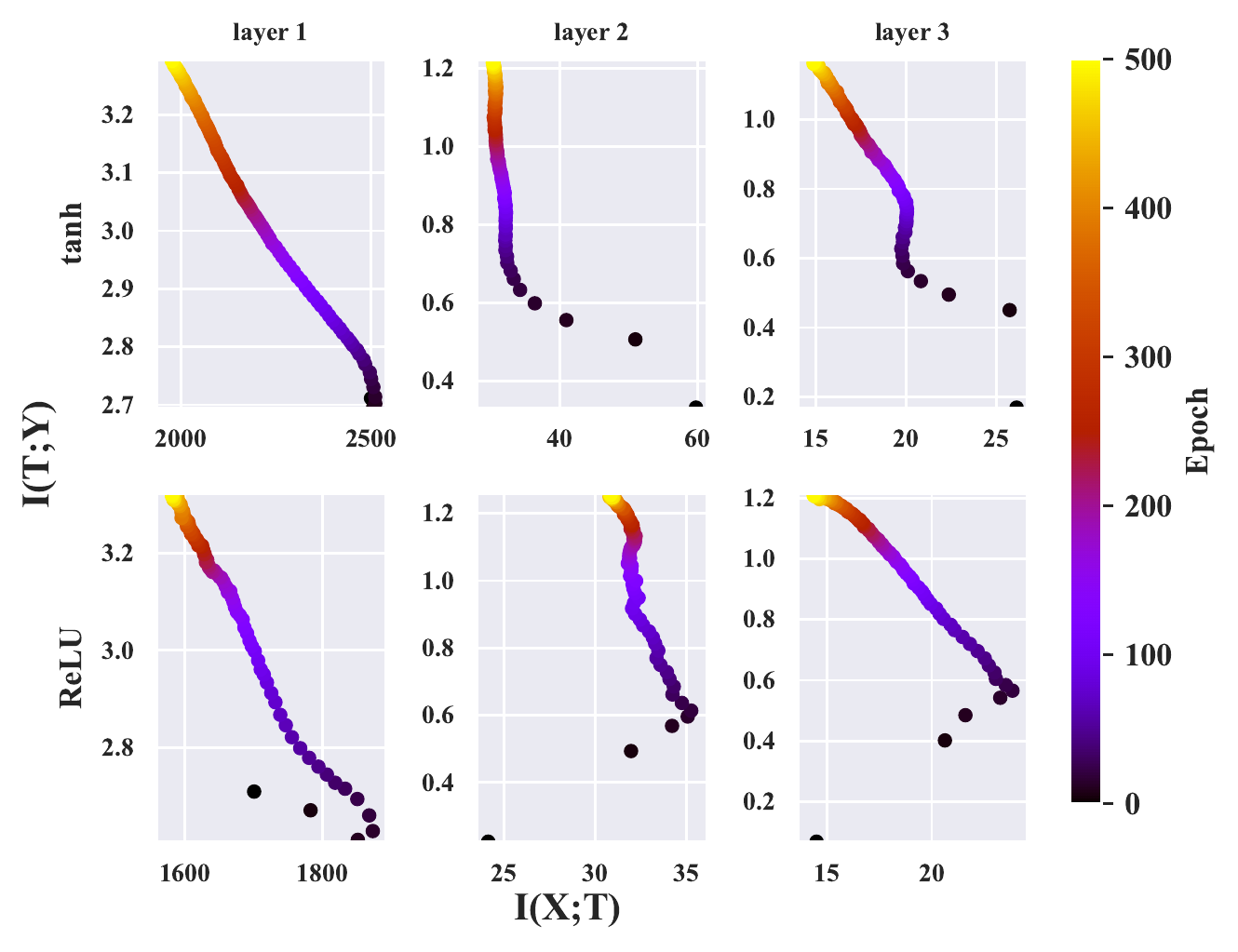}
}
\subfloat[IP of the 5-layer FCN]{
\includegraphics[width=0.56\textwidth]{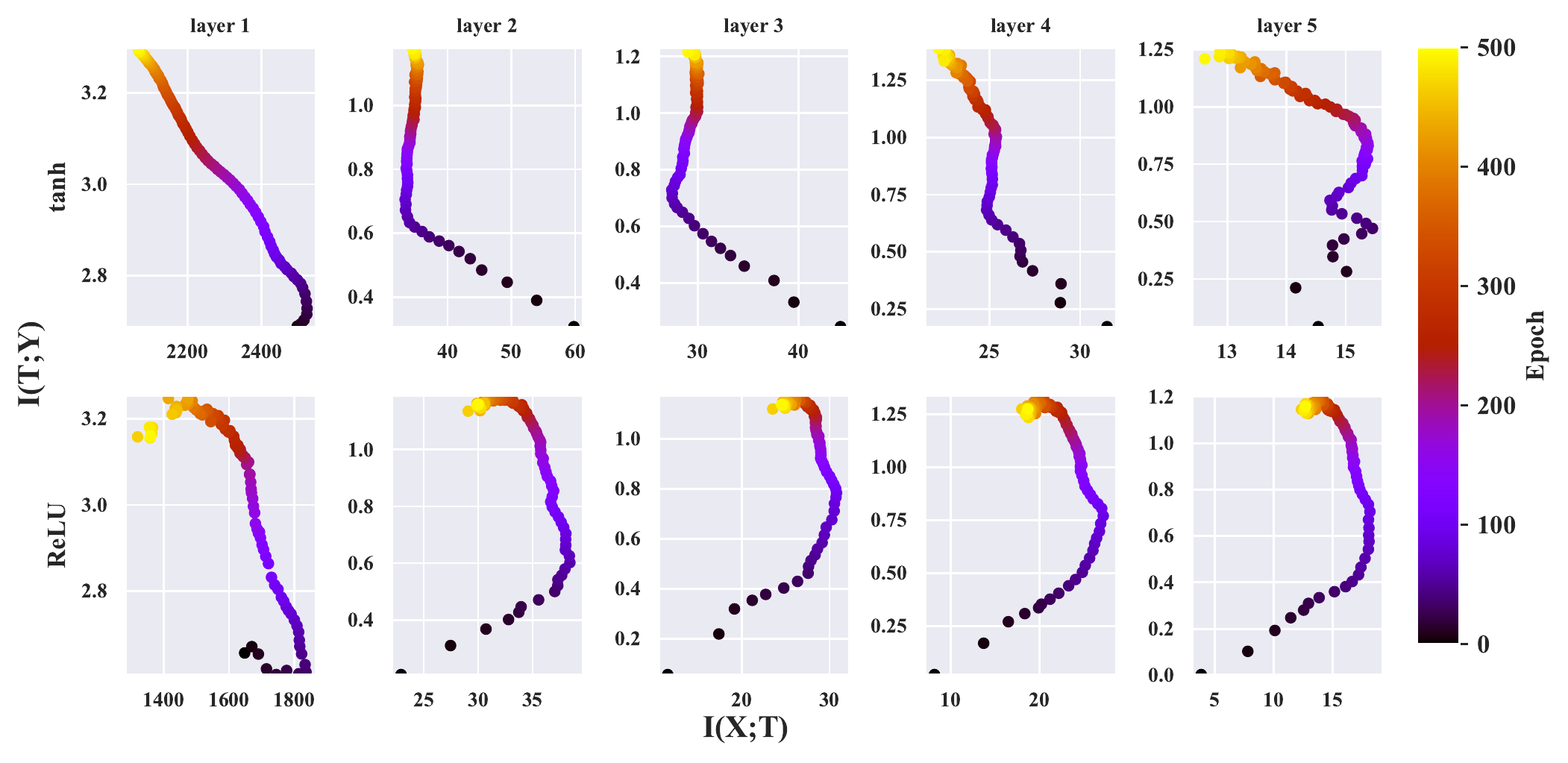}
}\\
\subfloat[IP of the 7-layer FCN]{
\includegraphics[width=0.8\textwidth]{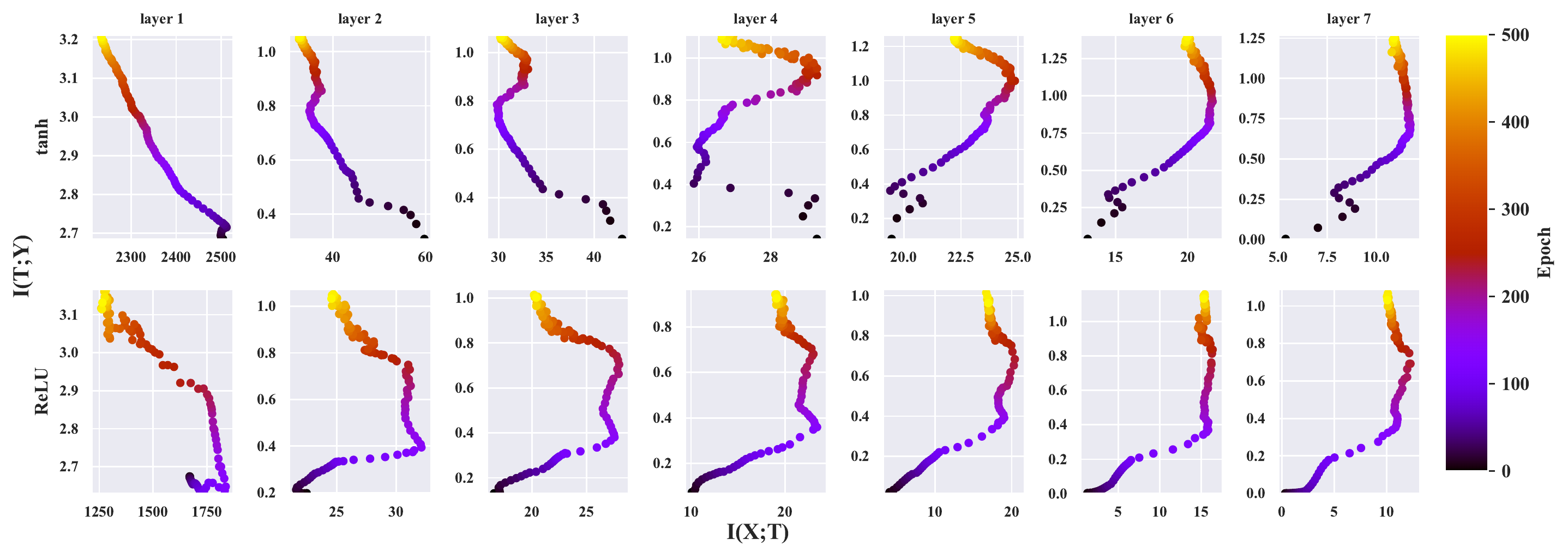}
}
\caption{Information Plane of fully connected neural networks with 3072-1024 in the first two layers and 10 in the last layer. Five and seven layers FCNs have multiple 20-neurons Middle layers, with 2 and 4 layers respectively. All models are trained on CIFAR10 dataset. Mutual information is estimated with 3000 test samples.}
\label{fnn_layer}
\end{figure*}

In this part, we employ LogDet estimator in deep neural network's behaviour analysis. Firstly, we display the information plane (IP) of a fully connected neural network activated by $tanh$ and $ReLU$. Our results find the compression in $I(X;T)$ occur in both $tanh$ and $ReLU$ activated networks. Then we compare the mutual information $I(X;T)$ to the introduced Layer Transmission Capacity $LTC(L)$ that measures the amount of information transmitted through each layer. Based on this, we try to give a plausible understanding of compression. Results reveal IB compression in $I(X;T)$ are caused by the reduction of Transmission Capacity that only exist in the first few layers. The Networks express two distinct LTC behaviour between shallow layers and deeper ones.

As we mentioned in the introduction section, to avoid unreliable results caused by misusing the estimator, we control the network conditions (models, activation functions, feature dimensions etc.) that LogDet estimator is proved effective in the above experiments. So we choose a fully connected neural network, with (784/3072-1024-20-20-10) neurons for each layer. Except for the input, the dimension of hidden layers is about up to 1000, which guarantees LogDet works well with 3000 samples. Networks are trained on benchmarks MNIST and CIFAR10, activated by $tanh$ and $ReLU$.

Firstly, we display the IP on the 5-layers FCN in Fig. \ref{fnn_rep}. The results show the phase transition in most of the cases, where $I(X;T)$ increasing at the beginning then decreasing. In Fig. \ref{fnn_rep} (a) that trained with CIFAR10, it shows a more sophisticated behaviour than (b) on MNIST, but their tendencies is roughly similar. Comparing with different activation function, we see the model with $tanh$ activation has rather a severe compression than operated by $ReLU$. This observation objects the findings in \cite{saxe2019information}, which account compression for the effect of $tanh$'s saturation and proposed compression doesn't exist in with $ReLU$. Our results show compression still occur in $ReLU$ where saturation is absent. Considering it is less severe than $tanh$, we also inference that the double-saturating $tanh$ can cause compression, but it is not the only reason. Then, to investigate whether such behaviour appears generally, we alter numbers of 20-neurons layer change the model to 3 and 7 layers. All 3, 5 and 7 layers' models are trained on CIFAR10. Our results displayed in Fig. \ref{fnn_layer} show deeper models tend to have stronger compression than shallow ones. Five and seven layers models show typical IB phase transition from fitting to compression. In each model, the former layer compresses more information about $X$ than the latter ones.

\begin{figure*}
\centering
\subfloat[LTC of the 3-layer FCN]{
\includegraphics[width=0.36\textwidth]{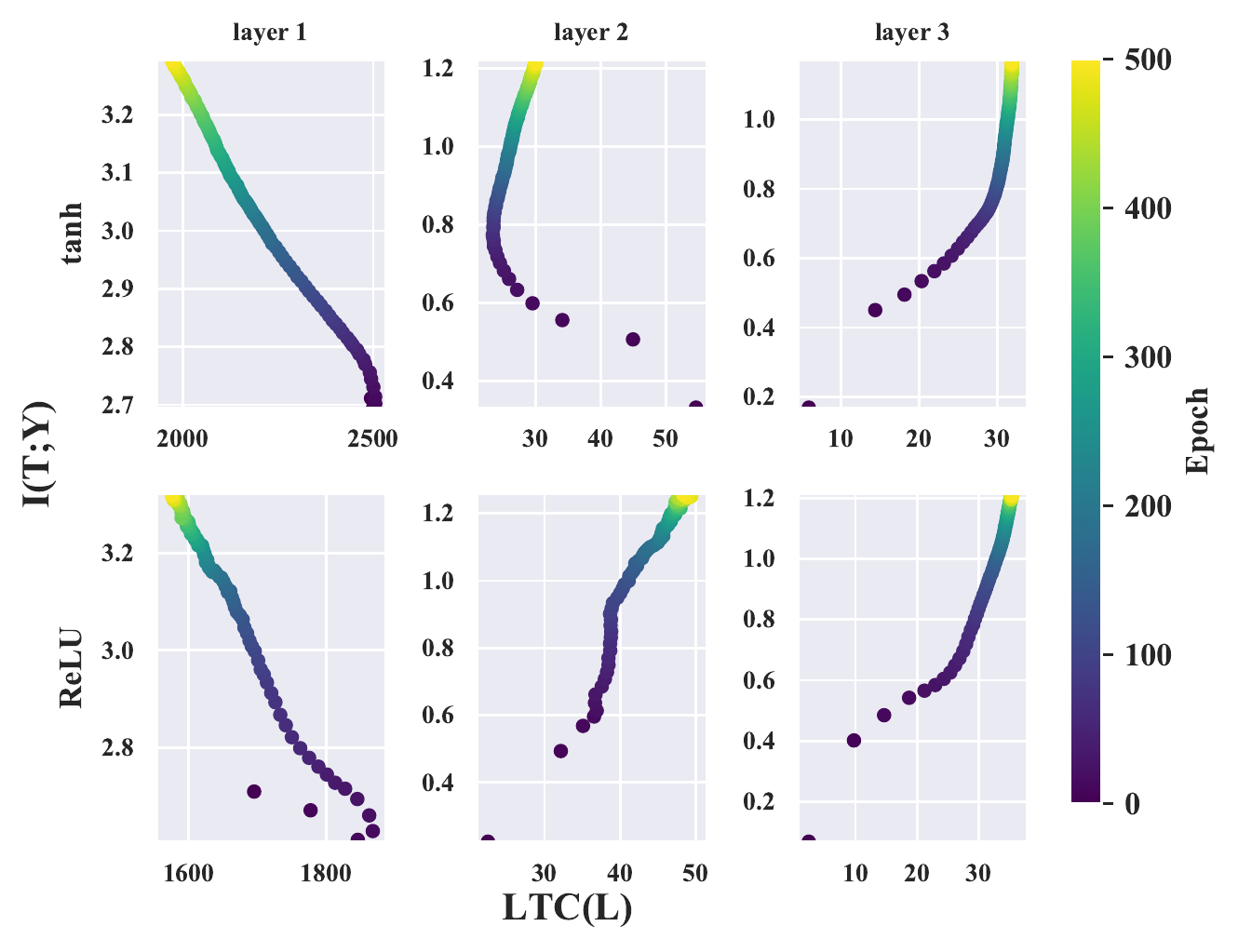}
}
\subfloat[LTC of the 5-layer FCN]{
\includegraphics[width=0.56\textwidth]{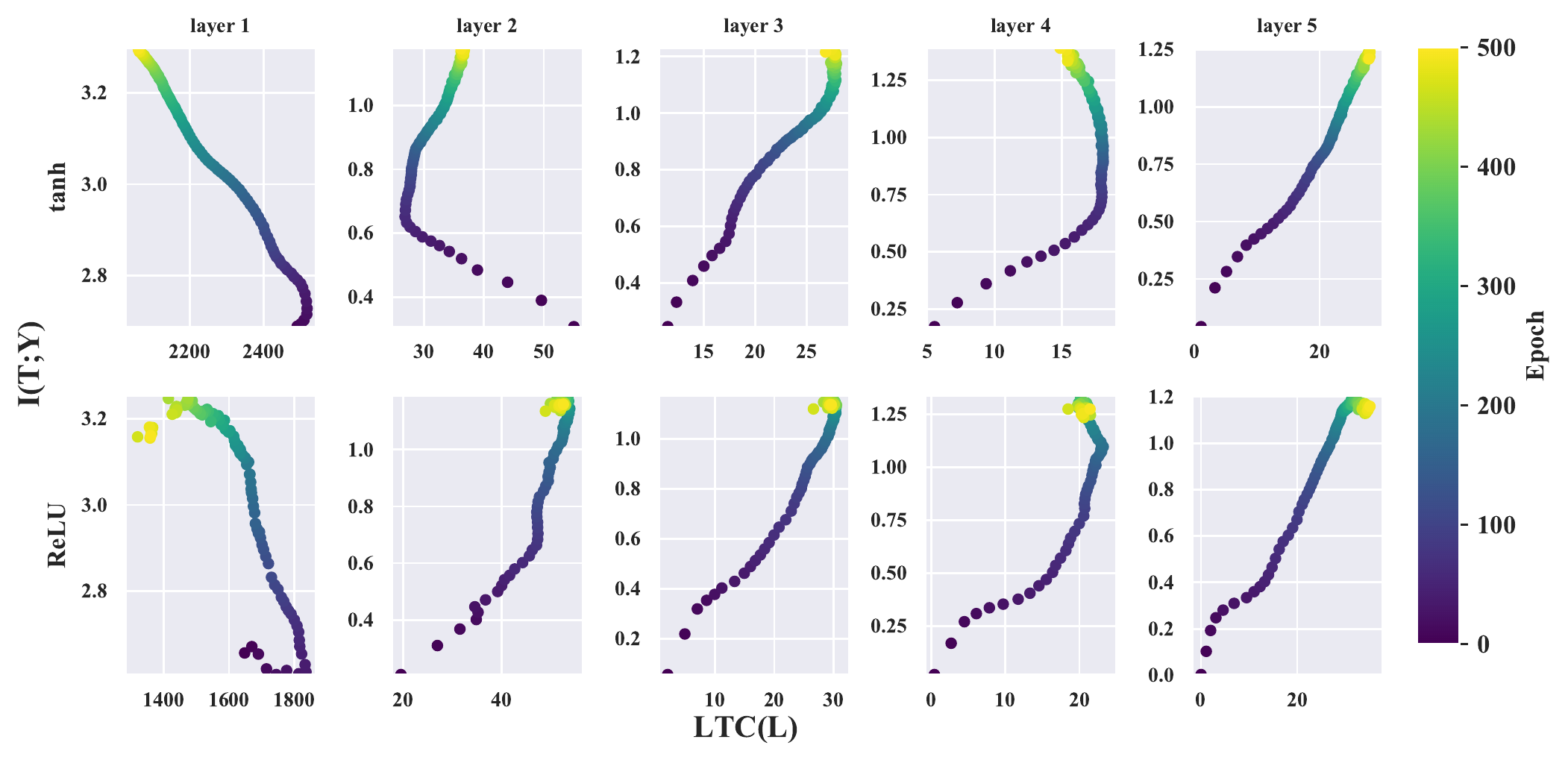}
}\\
\subfloat[LTC of the 7-layer FCN]{
\includegraphics[width=0.8\textwidth]{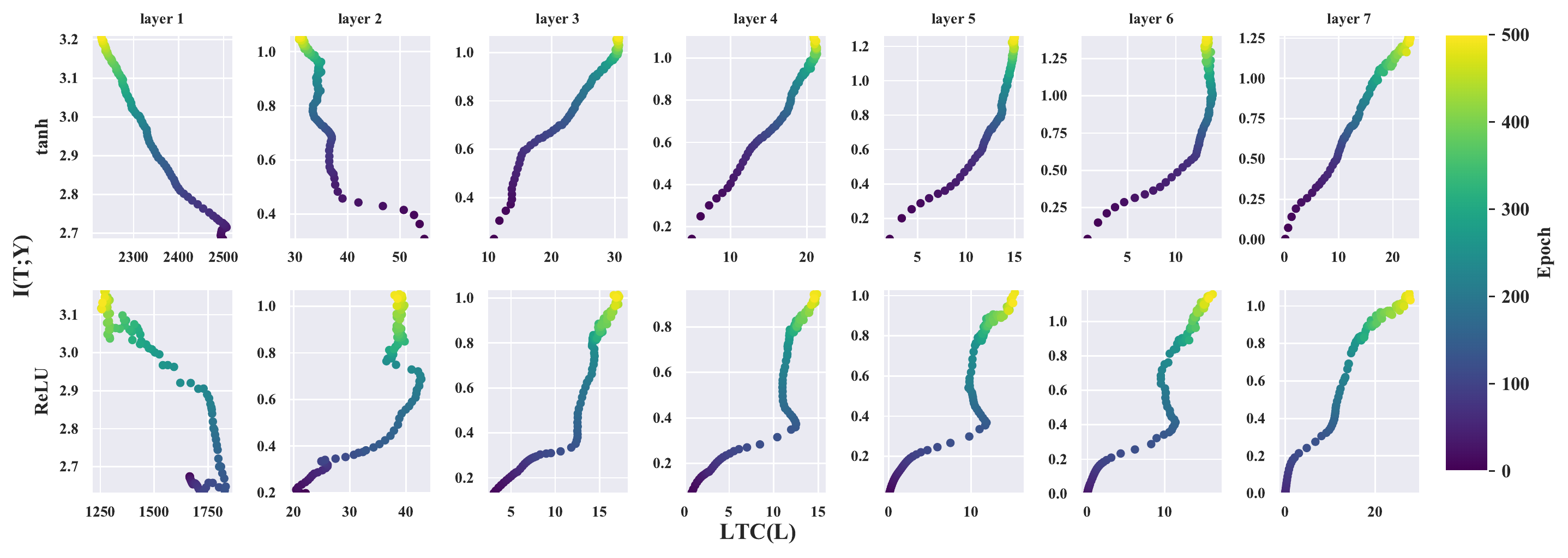}
}
\caption{Layer Transmission Capacity of different layers' fully connected neural networks that possess same structure as the ones in information plane demonstration. All models are trained on CIFAR10 dataset and estimated with 3000 test samples.}
\label{fnn_Capacity}
\end{figure*}

However, because each layer's $I(X;T)$ behaviour is affected by all its former layers, only observing such information plane cannot reveal precisely the detailed behaviour of the neural network. To further understand the compression phenomenon, we estimate Layer Transmission Capacity and construct a layer-wise information plane, where the horizontal axis $I(X;T_i)$ is replaced by Layer Transmission Capacity $LTC(L_i)$, which indicates how much information is transmitted in $ith$ layer. The vertical axis of $I(T_i;Y)$ is kept and interpreted as the amount of target-related information that passes the $L_i$ layer. In Fig.\ref{fnn_Capacity}, we display the layer-wise Information Planes of the three corresponding FCNs in Fig.\ref{fnn_layer}.

In all six results (with different activation and model depth), we see a decrease of $LTC$ in the first few layers. This decrease directly tells the first layer of FCN reduces the amount of information transmitted through it. Except for the decrease in the first layer(and to some $tanh$ activated situation in the second layer), there is hardly any reduction of $LTC$ observed. Most later layers' $LTC$ keeps increasing, indicating the layers are expanding their channel and allowing more information to pass by. Thus compression never happens in their layer-wise demonstration. We argue that this result doesn't contradict with the compression observed in IB plane, but rather gives an interpretation: The input layer reduces the information about $X$ goes into deeper layers. Despite other layers keep increasing their transmission capacity, the $I(X;T_i)$ will still decrease for it is limited in the beginning. In which case, the decrease of $I(X;T)$ displayed in Fig.\ref{fnn_layer} is mainly an influence of compression in the first layer.

\begin{figure*}
\centering
\subfloat[IP with fixed first layer]{
\includegraphics[width=0.49\textwidth]{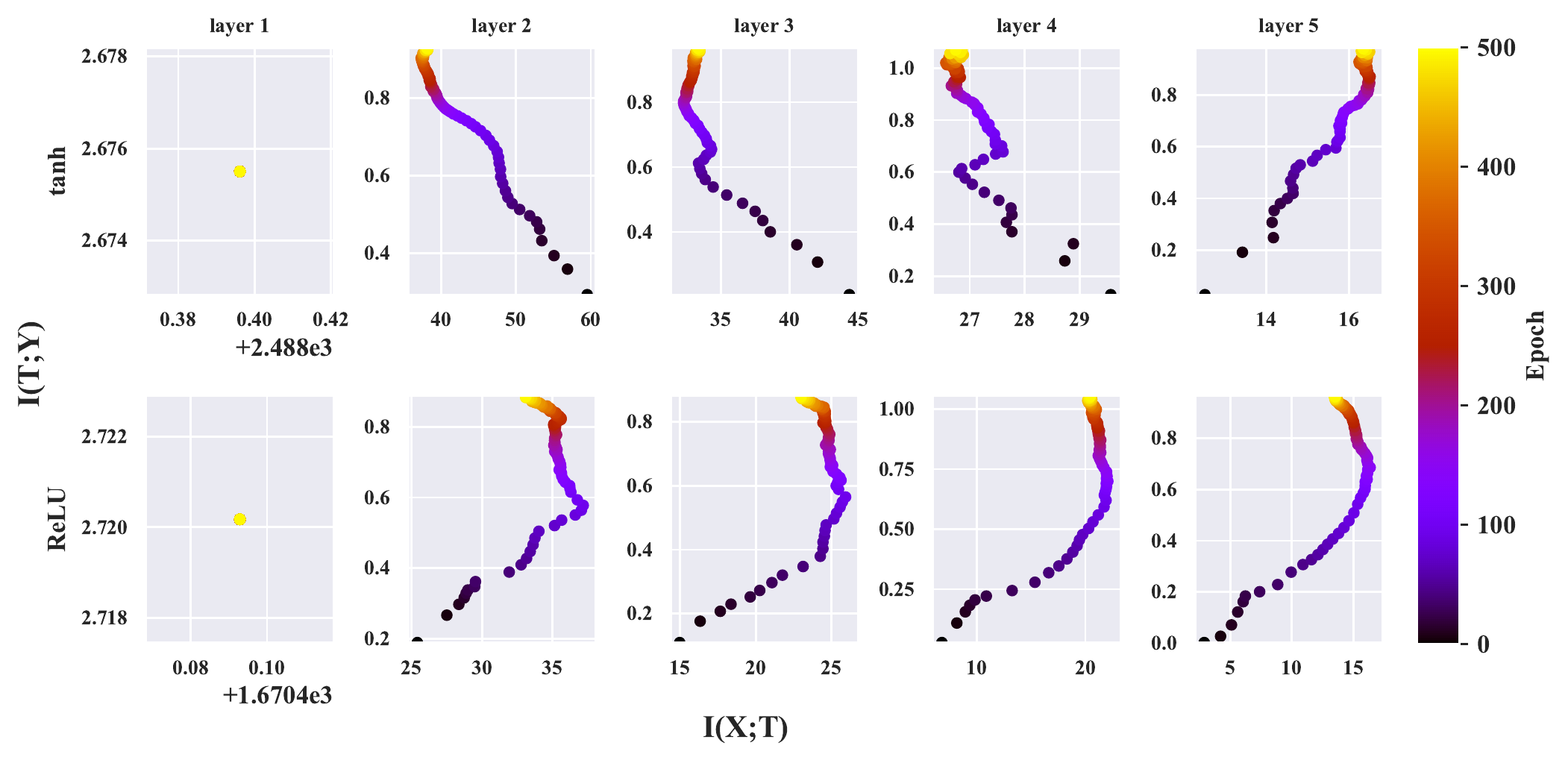}
}
\subfloat[IP with fixed second layer]{
\includegraphics[width=0.49\textwidth]{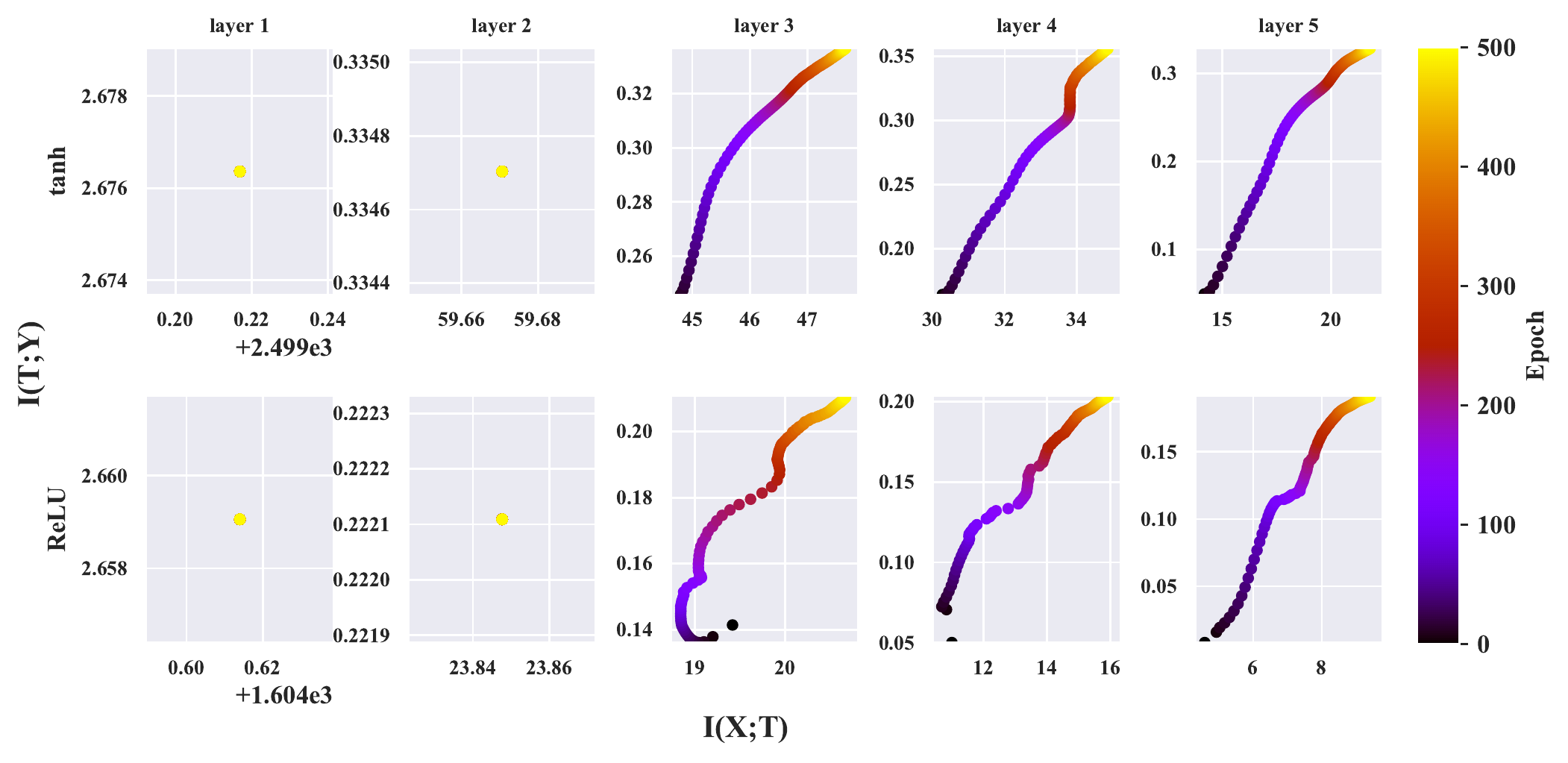}
}
\hfil
\subfloat[LTC with fixed first layer]{
\includegraphics[width=0.49\textwidth]{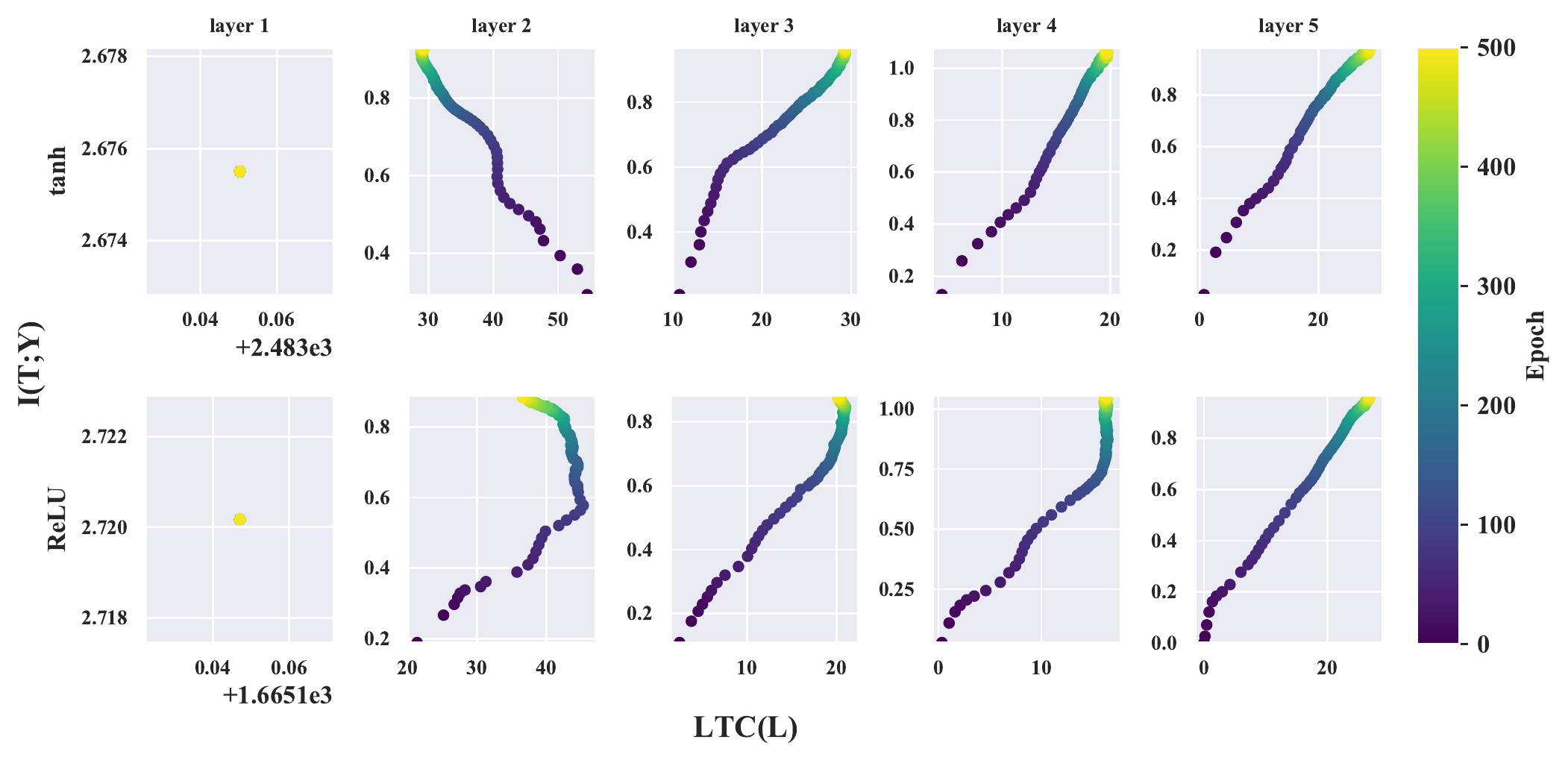}
}
\subfloat[LTC with fixed second layer]{
\includegraphics[width=0.49\textwidth]{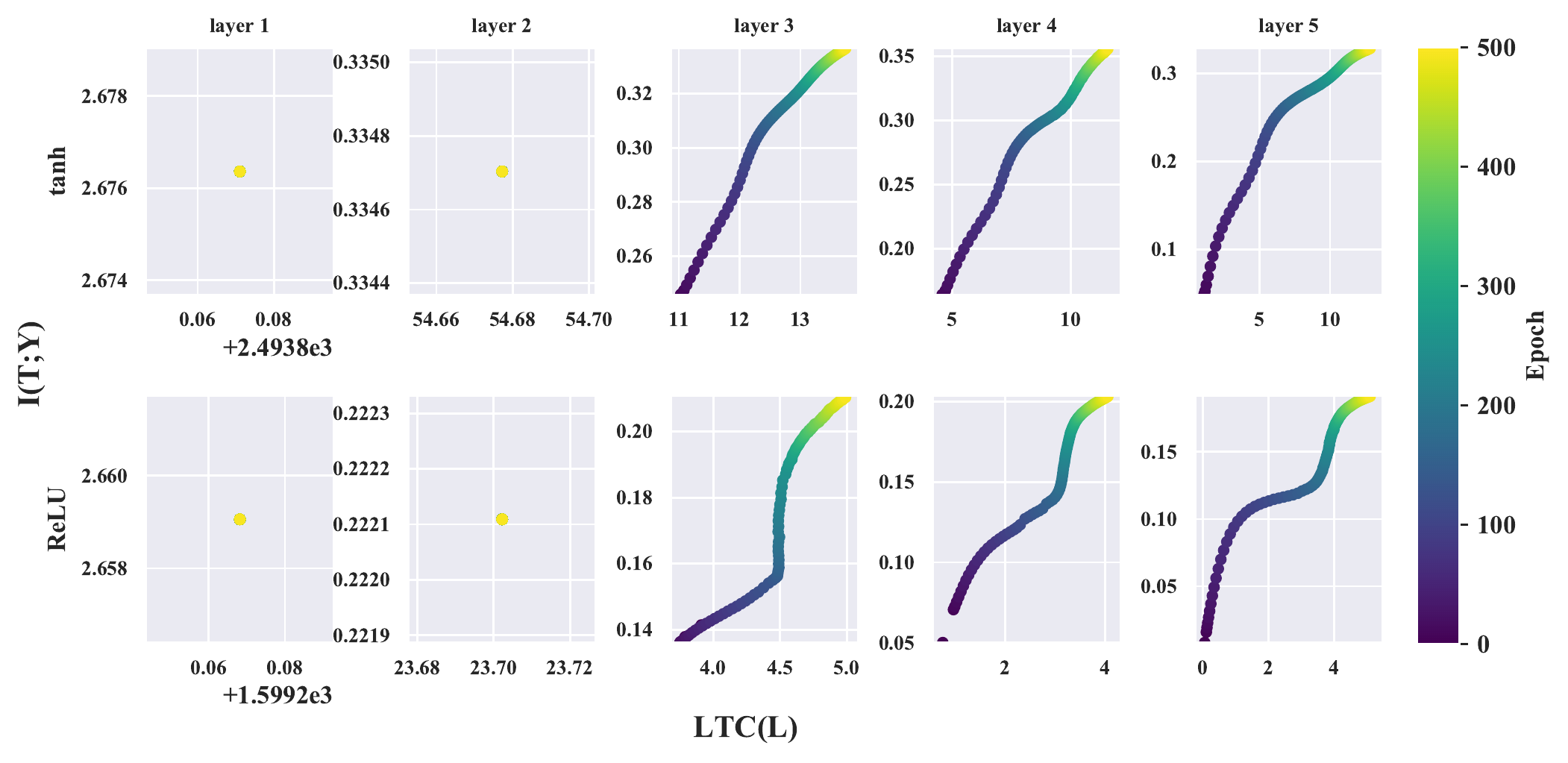}
}
\hfil
\subfloat[ACC with fixed first layer]{
\includegraphics[width=0.35\textwidth]{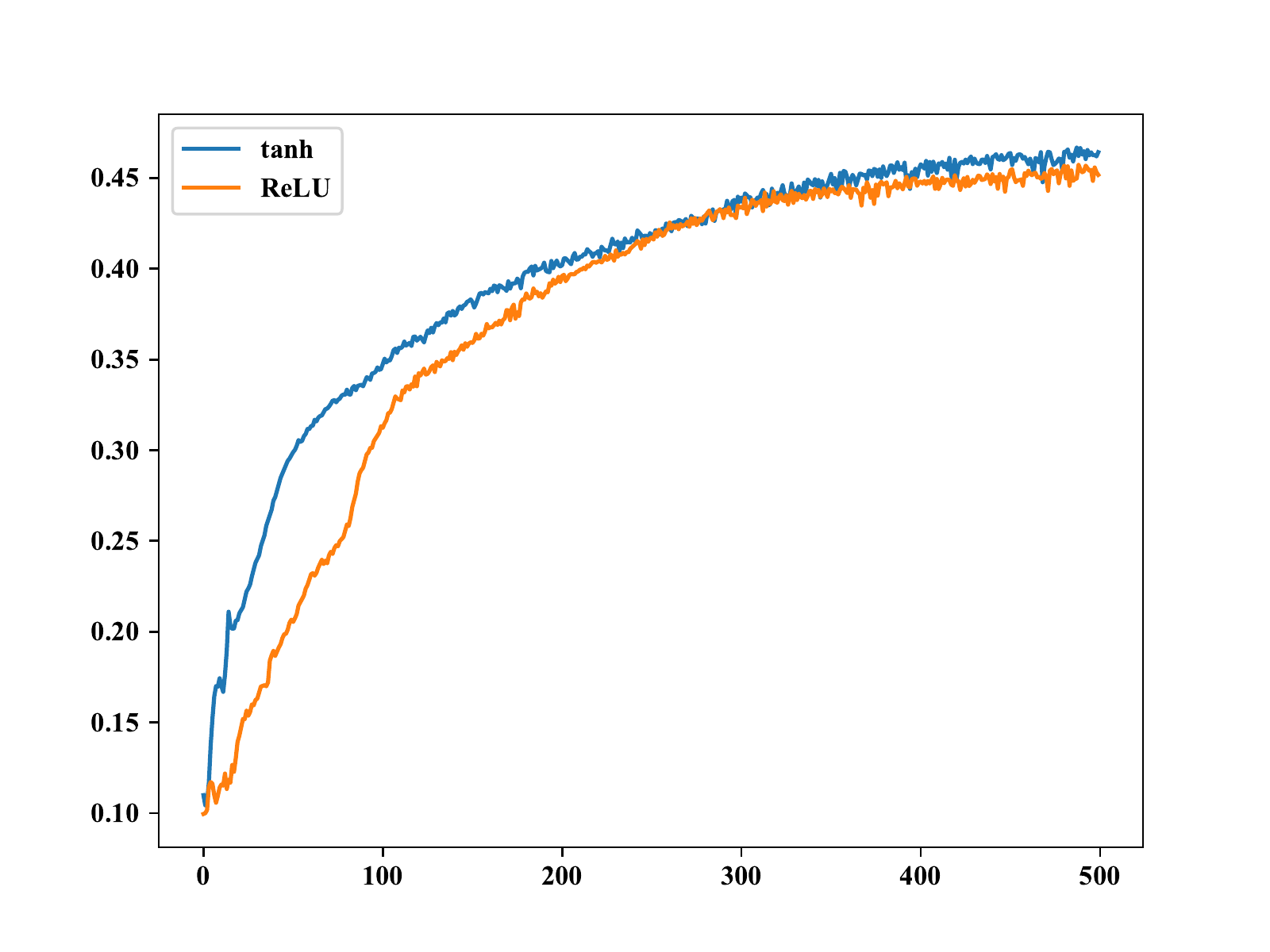}
}
\subfloat[ACC with fixed second layer]{
\includegraphics[width=0.35\textwidth]{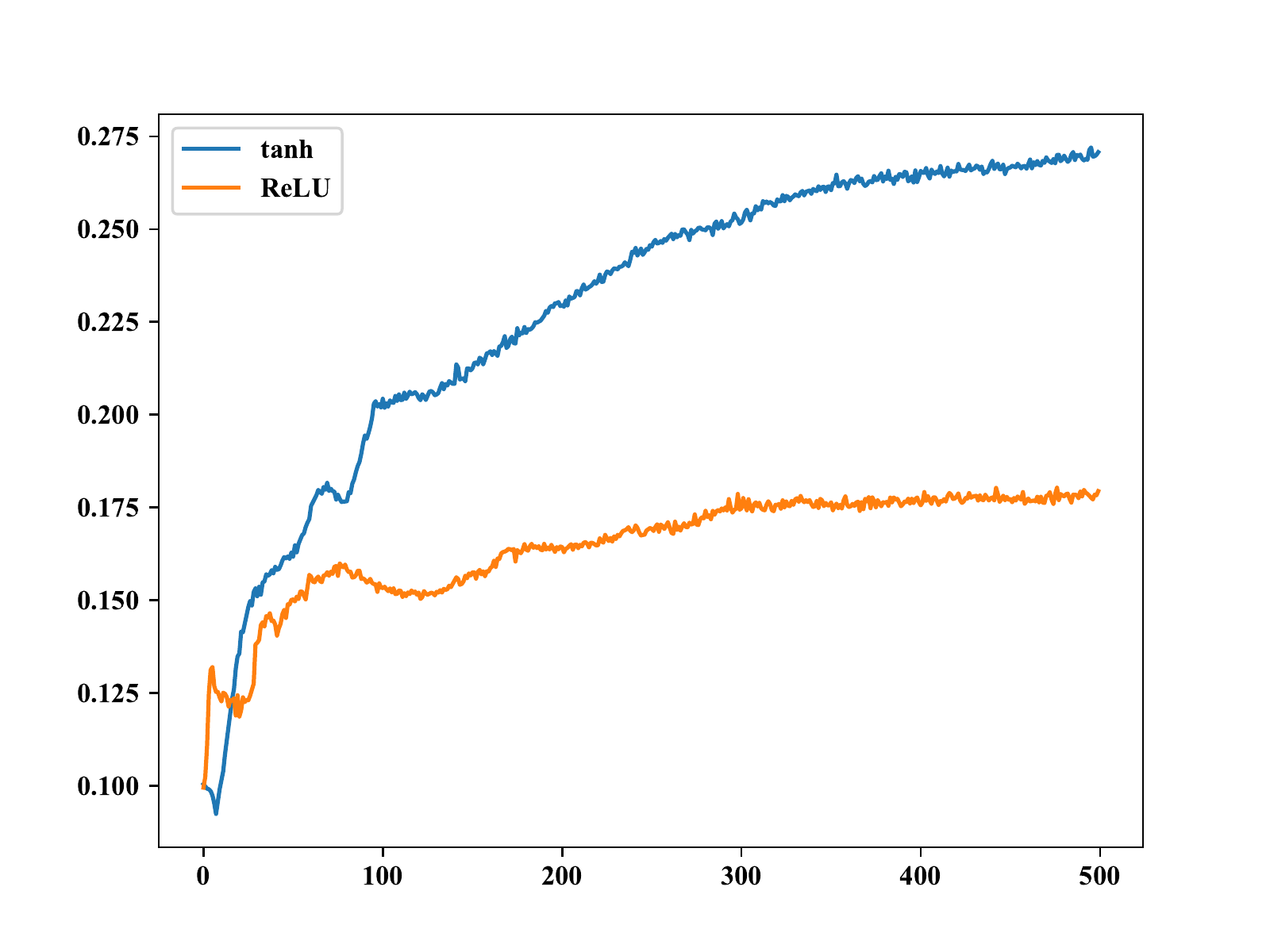}
}
\caption{Information Plane, Layer Transmission Capacity and Accuracy of the 5-layer FCN with fixed parameters of first layer (a,c,e) and fixed first and second layers (b,d,f). Trained on CIFAR10 dataset, the model compresses in second layer of LTC when fixed the first layer. Where such compression is absent without parameter fixing.}
\label{fnn_essential_1}
\end{figure*}

Also worth noticing is, the latter layers tend to have similar behaviour that is distinct to the former ones. This reveals a functional distinction along with the depth of neural network, where the former layer tends to compress the data, acting like feature selector to get a useful representation. While the latter ones are optimized to transmit information more effectively or change its representation to fit the target. Such phenomenon that appears in three models with different depth shows the generality of LTC behaviour. Furthermore, we want to investigate whether such phenomenon is essential, we fix the first layer's parameters, then train the network as usual. If such two-step compression-transmission phenomenon is essential to training neutral network, we expect compression of LTC would appear in the second layer where it doesn't promised appear in former test. The result is displayed in Fig.\ref{fnn_essential_1}. We see the first layer is fixed as a stable point. Surprisingly, the second layer, that is not compressed in the former experiment, showing a similar $LTC$ compression as the first layer. The Compression-Transmission distinction happens again! This further support the essences of such compression behaviour of the first layer. Also interesting is, when we fix more parameters that the model can be hardly trained on the task, such compression in LTC disappeared. Along with its absence, the prediction accuracy can be merely improved through training. This observation connected the compression of the first layer to the model's performance, and again shows first layer's compression is essential
in training neural networks.

In all, these results explain the cause of IB compression as the effect of the first layer. It also reveals the distinct behaviour between the first few layers and other layers. We show such distinction is a general phenomenon when training neural networks by changing the Network structures, activation functions and fixing parameters, and such phenomenon is connected with prediction performance.

% \begin{figure*}
% \centering
% \subfloat[IP]{
% \includegraphics[width=0.35\textwidth]{zhou9_1.pdf}
% }
% \subfloat[CL]{
% \includegraphics[width=0.35\textwidth]{zhou9_2.pdf}
% }
% \subfloat[ACC]{
% \includegraphics[width=0.25\textwidth]{zhou9_3.pdf}
% }
% \caption{Information Plane, Layer Transmission Capacity and Accuracy of the 5-layer FCN with fixed parameters of first and second layers.}
% \label{fnn_essential_2}
% \end{figure*}

\section{Discussion and Future Works}

By approximating Shannon Differential Entropy, this work shows LogDet estimator is capable of estimating information entropy on artificial samples and real-world benchmarks. Experiments show LogDet estimator can be applied to any high dimensional features and is robust to random noise. Comparing to other analytical works, these tests ensure the soundness of informative behaviour analysis of neural networks with LogDet estimator.

Furthermore, in neural networks' IB analysis, our work gives a layer-wise interpretation for compression behaviour. First we find networks experience compression in both double and single saturating activation. Experiments reveal saturation of $tanh$ can cause compression. This explains why we often witness more severe compression in $tanh$ activated network. Also, compression occurs in $ReLU$ activated FCNs proves the double-saturating effect is not the only reason. To better understand such phenomenon, We simplify the discussion by a layer-wise Transmission Capacity analysis. The result shows only the first or second layer reduces its transmission capacity LTC, causing the decrease of $I(X;T)$ as compression phenomenon observed in IB plane. Besides, the model learns to automatically divide into two areas: \textbf{(1)} shallow layers transmit the input related information selectively; \textbf{(2)} deeper layers on the opposite, keep increasing their capacity to avoid transmitting loss when processing the features. These results reveal a functional difference among neural network layers which gives an empirical guideline for future analysis of neural networks informative behaviour.

We believe LogDet estimator provides a extendable framework for estimating informative relations and constructing informational objectives. Some potential directions are displayed below:
\begin{enumerate}
    \item\textbf{Estimating covariance matrix}: In this work, LogDet estimator measures entropy with covariance matrix. In which case, we can adopt many covariance estimation methods to improve. One possible approach is to introduce prior knowledge about estimated samples. For example, when estimating with image data, the locally correlated nature of pixels can be introduced. Therefore, we can reduce covariance between distant variables, while keep only the correlation of their $ith$ neighbours. This can be achieved by eliminating other but the diagonal and off-diagonal values in covariance matrix $\Sigma$. Covariance estimation has been well studied in modern statistics, we refer to \cite{hu2017comparison}\cite{lam2020high} for potential improvements.
    \item \textbf{Kernel extension}: Since the multi-signal extension reveals the potential of kernelizing the similarity matrix $\Sigma$ in $\log\det(I+\beta\Sigma)$, we can replace it with Gram matrices where each component is an arbitrary kernel function $k(i,j)$. Similar kernel extension using $\log\det$ function has been applied in \cite{wright2007classification}\cite{yu2021deep} and is shown flexible in visual recognition problems. We believe this can also be applied to enhance the LogDet information measuring on more sophisticated sample distribution.
    \item \textbf{Learning Objectives}: LogDet expression has revealed the potential to act as an objective in supervised learning tasks, and express robust performance to corrupted labels\cite{yu2020learning}. By further analysing the network behaviour, it is possible to use LogDet as objective function and optimize it through backpropagation. For example, we tried to use a kernel function $exp(x^Ty-1)$ to force the similarity matrix to only captures positive correlations. Then, we use $H_D(Y|T)=H_D(T,Y)-H_D(T)$ as a learning objective. This expression can be used to training neural network classifiers, and can increase accuracy through training. However, since $exp$ cannot eliminate all negative correlations while retaining the positive ones, our objective is not an ideal adaptation. We believe this is a quite promising direction.
\end{enumerate}
We also witness the limitations of LogDet estimator. As a matrix-based method, it heavily relies on the amount of estimating samples. Although the approximating method avoids the singular problem, the estimating result is still discredited when the sample number is less than the feature dimension. This currently is an unavoidable problem of matrix-based methods. Similar limitations also occur in the recent proposed $\alpha$-$R\acute{e}nyi$ estimator. Despite it proposed independent with feature dimension, its estimating result is less accurate when sample amount is fewer than feature dimension(see Appendix. B for details test). We look forward to further improvements to overcome such limitation.

% or


\section*{Acknowledgment}
This work is supported by Natural Science Foundation of Tianjin (20JCYBJC00500),
the Science \& Technology Development Fund of Tianjin Education Commission for Higher Education (2018KJ217), the Tianjin Science and Technology Program (19PTZWHZ00020) and Natural Science Foundation of China (61772363).

\ifCLASSOPTIONcaptionsoff
  \newpage
\fi



%\bibliographystyle{IEEEtran}
\bibliography{arxiv.bib}


% \end{IEEEbiography}


% % insert where needed to balance the two columns on the last page with
% %\newpage



%\vfill


\appendices
\section{Proof of main results}
$Lemma$ 1 To any positive definite matrix $A$ and positive semi-definite matrix $B$, we have the additive inequality:
\begin{equation*}
    \det(A+B)\geq{\det(A)}
\end{equation*}
    
$Proof$ since A is an arbitrary positive definite matrix, there exist a invertable positive semi-definite matrix T where $T^2=A$. Then, we have:
\begin{align*}
    \det(A+B)&=\det(TIT+B)\\
    &=\det(T(I+T^{-1}BT^{-1})T)\\
    &=\det(T)\det(T)\det(I+T^{-1}BT^{-1})\\
    &=\det(A)\det(I+T^{-1}BT^{-1})
\end{align*}
Let $Q=T^{-1}BT^{-1}$, and eigenvalues of $Q$ as $[\lambda_i]_{i=1}^n$, then Q is still positive semi-definite, now we have:
\begin{align*}
    \det(A)\det(I+T^{-1}BT^{-1})&= \det(A)\det(I+Q)\\
    &=\det(A)\prod_{i=1}^n(1+\lambda_i)\\
    &\geq{\det(A)(1+\prod_{i=1}^n\lambda_i})\\
    &=\det(A)(1+\det(Q))\\
    &\geq{\det(A)}
\end{align*}
The last inequality is for $\det(Q)\geq{0}$ as it is positive semi-definite.

$Commutative$ $property$
\begin{align*}
    \frac{1}{2}\log\det(I+\beta\frac{X^TX}{n})=\frac{1}{2}\log\det(I+\beta\frac{XX^T}{n})
\end{align*}

$Proof$ of Eq.\ref{eq4}:
performing $SVD$ to $X$, we have $X=UDV$ where $D=diag([n\lambda_1,n\lambda_2,...,n\lambda_k])$ that $\lambda_i$ is $\frac{X^TX}{n}$'s ith eigenvalue, then:
\begin{align*}
    \mathbf{H}_D(X)&=\frac{1}{2}\log\det(I+\beta\frac{X^TX}{n})\\
    &=\frac{1}{2}\log\det(I+\frac{\beta}{n}V^TD^2V)\\
    &=\frac{1}{2}\log\det(V^T(I+\frac{\beta}{n}D^2)V)\\
    &=\frac{1}{2}\log\det(I+\frac{\beta}{n}D^2)\\
    &=\frac{1}{2}\log\prod_{i=1}^k(1+\beta\lambda_i)\\
    &=\frac{1}{2}\sum_{i=1}^k\log(1+\beta\lambda_i)\\
    &=\frac{1}{2}\sum_{i=1}^k\log(\frac{1}{\beta}+\lambda_i)+\frac{k}{2}\log\beta
\end{align*}

$Proof$ of $Proposition$ 2.1: 
The first inequality $\mathbf{H}_D(X_1,X_2)\leq\mathbf{H}_D(X_1)+\mathbf{H}_D(X_2)$ 
Let estimated covariance matrix $Cov(X_1)=\Sigma_1=\frac{X_1^TX_1}{n}$, $Cov(X_2)=\Sigma_2=\frac{X_2^TX_2}{n}$, and $Z=[X_1,X_2]$, $Cov(Z)=\begin{pmatrix} \Sigma_1 & F \\ F^T & \Sigma_2 \end{pmatrix}$, we have
\begin{align*}
    \mathbf{H}_D(X_1)+\mathbf{H}_D(X_2)&=\frac{1}{2}\log\det(I+\beta\Sigma_{X_1})\\
    &+\frac{1}{2}\log\det(I+\beta\Sigma_{X_2})\\
    &=\frac{1}{2}\log\det(I+\beta\frac{X_1^TX_1}{n})(I+\beta\frac{X_2^TX_2}{n})\\
    &=\frac{1}{2}\log\det(I+\beta\frac{X_1X_1^T}{n})(I+\beta\frac{X_2X_2^T}{n})\\
    &=\frac{1}{2}\log\det(I+\beta\frac{X_1X_1^T}{n}+\beta\frac{X_2X_2^T}{n}\\
    &+\beta^2\frac{X_1X_1^T+X_2X_2^T}{n^2})\\
    &\geq{\frac{1}{2}\log\det(I+\beta\frac{X_1X_1^T}{n}+\beta\frac{X_2X_2^T}{n})}\\
    &=\frac{1}{2}\log\det(I+\beta\frac{ZZ^T}{n})\\
    &=\frac{1}{2}\log\det(I+\beta\frac{Z^TZ}{n})\\
    &=\mathbf{H}_D(X_1,X_2)
\end{align*}
The last inequality satisfied for $Lemma$ 1.
The second inequality $\mathbf{H}_D(X_1,X_2)\geq\max(\mathbf{H}_D(X_1),\mathbf{H}_D(X_2))$ can be derived using above expression:
\begin{align*}
    \mathbf{H}_D(X_1,X_2)&=\frac{1}{2}\log\det(I+\beta\frac{Z^TZ}{n})\\
    &=\frac{1}{2}\log\det(I+\beta\frac{X_1X_1^T}{n}+\beta\frac{X_2X_2^T}{n})\\
\end{align*}
Applying $Lemma$ 1, we have:
\begin{align*}
    \mathbf{H}_D(X_1,X_2)&\geq{\frac{1}{2}\log\det(I+\beta\frac{X_1X_1^T}{n})}= \mathbf{H}_D(X_1)\\
    \mathbf{H}_D(X_1,X_2)&\geq{\frac{1}{2}\log\det(I+\beta\frac{X_2X_2^T}{n})}= \mathbf{H}_D(X_2)
\end{align*}
So, $$\mathbf{H}_D(X_1,X_2)\geq{\max(\mathbf{H}_D(X_1),\mathbf{H}_D(X_2))}$$

$Proof$ to $Corollary$ 4.1: The procedure is similar with the proof to Proposition 2.1. By applying commutative property of $\log\det$ function and $Lemma$ 1, the inequality is easy to prove.

\section{Limitations of Matrix-based Estimators}
The matrix-based method heavily relies on the amount of estimating samples. Although, with samples far less than the dimension, LogDet is express the capacity to show the relations in $I_D(X;Y)$ Fig. \ref{MI}. However, Gaussian samples are too simple, the subtle dependence change in neural networks requires more samples to reveal. We conduct the activation function's experiment to demonstrate:
\begin{figure*}
\centering
\subfloat[fixed amount of samples while increase $d$]{
\includegraphics[width=0.8\textwidth]{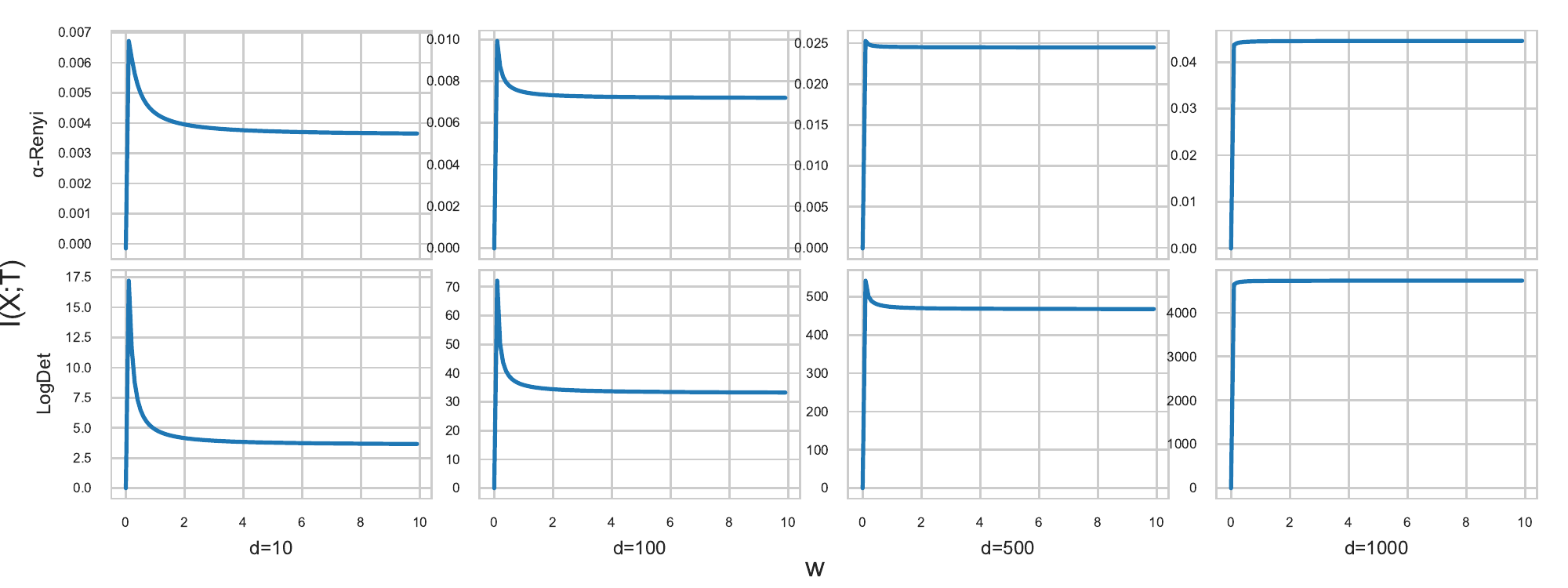}
}
\hfil
\subfloat[fixed $d$ while increase the amount of samples]{
\includegraphics[width=0.6\textwidth]{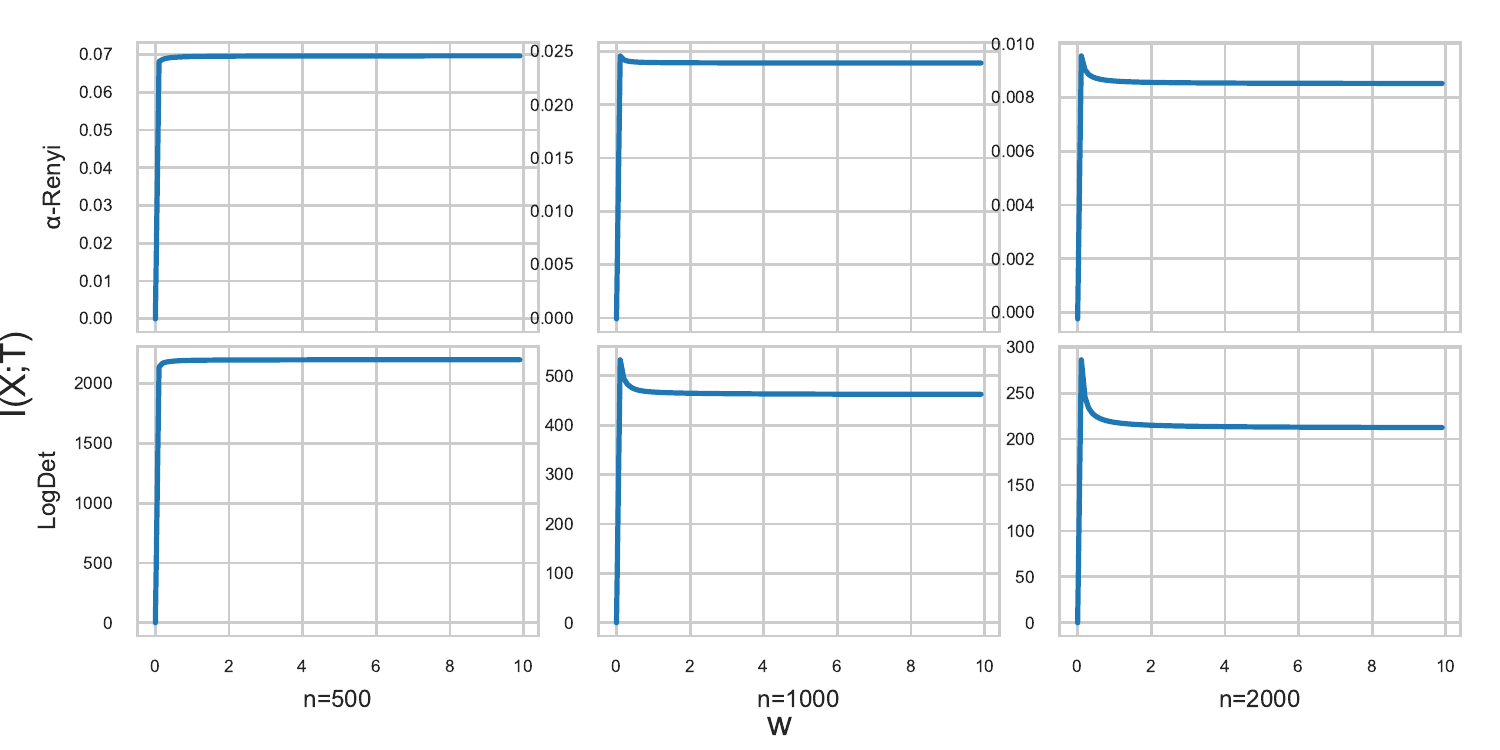}
}
\caption{Limitation of matrix-based methods using $tanh$'s saturating test.}
\label{zhou10}
\end{figure*}

In Fig. \ref{zhou10}(a), we compare the LogDet estimator and matrix-based $\alpha$-$R\acute{e}nyi$ estimator on $tanh$'s saturation effect, and use 500 samples to estimate. It is clear, when the feature dimension increases, the saturation effect is shown by both methods become less obvious. At last it disappear when $d=1000$. But when we fix the dimension as 1000, and increase the sample numbers from 500 to 2000 in (b), such decrease a.k.a. compression shows again. We know such saturation exists, but when the dimension is larger than samples, cannot be revealed by both matrix-based estimators. Also interesting is that, in all results, LogDet estimator express more severe change than $\alpha$-$R\acute{e}nyi$ estimator. 

\end{document}